%% file: main.tex
\documentclass[10pt,twocolumn,letterpaper]{article}
\usepackage[T1]{fontenc}

\usepackage{wacv}                

\definecolor{wacvblue}{rgb}{0.21,0.49,0.74}
\usepackage[pagebackref,breaklinks,colorlinks,allcolors=wacvblue]{hyperref}

\usepackage[capitalize]{cleveref}

\input{preamble}

\usepackage{graphicx,subcaption} 
\usepackage{algorithm}
\usepackage{algpseudocode}
\usepackage{amsmath}
\usepackage{xcolor}
\usepackage{amssymb}
\usepackage{multirow}
\usepackage{booktabs}
\usepackage{pifont}
\usepackage{tabularx}

\def\wacvPaperID{349} 
\def\confName{WACV}
\def\confYear{2027}

\title{Aligning Misaligned Benchmarks for Single-View Novel View Synthesis}

\author{
  Yang-Che Tseng\textsuperscript{1}\quad
  Erqun Dong\textsuperscript{1}\quad
  Jeong-gi Kwak\textsuperscript{1}\quad
  Gopal Sharma\textsuperscript{1}\quad
  Shakiba Kheradmand\textsuperscript{1}\quad
  \\
  Stephan Garbin\textsuperscript{2}\quad
  Hossam Isack\textsuperscript{2}\quad
  Kwang Moo Yi\textsuperscript{1}
  \\
  \textsuperscript{1} University of British Columbia\quad
  \textsuperscript{2} Google
}

\begin{document}

\maketitle

\input{sections/0_abstract}
\input{sections/1_introduction}

\input{sections/2_related_works}
\input{sections/3_eval_protocol}

\input{sections/4_video_nvs}
\input{sections/5_results}
\input{sections/6_conclusion}

\section*{Acknowledgments}
This work was supported in part by the Natural Sciences and Engineering Research Council of Canada (NSERC) Discovery Grant, NSERC Alliance Advantage Grant, Google, Digital Research Alliance of Canada, and Advanced Research Computing at the University of British Columbia.

{
    \small
    \bibliographystyle{ieeenat_fullname}
    \bibliography{macros.bib, main}
}

\clearpage
\input{sections/X_supplementary}

\end{document}

%% file: preamble.tex
\usepackage{soul}

\usepackage{makecell}
\usepackage{booktabs}
\usepackage{array}
\usepackage{wrapfig}

\usepackage{enumitem}
\usepackage{cleveref}
\crefname{table}{Table}{Tables}
\Crefname{table}{Table}{Tables}
\crefname{figure}{Fig.}{Figs.}
\Crefname{figure}{Figure}{Figures}
\crefname{section}{Sec.}{Secs.}
\Crefname{section}{Section}{Sections}

\definecolor{chocolate}{rgb}{0.48, 0.25, 0.0}
\definecolor{bole}{rgb}{0.47, 0.27, 0.23}
\definecolor{goldenpoppy}{rgb}{0.99, 0.76, 0.0}
\definecolor{goldenyellow}{rgb}{1.0, 0.87, 0.0}
\definecolor{palesilver}{rgb}{0.79, 0.75, 0.73}
\definecolor{bronze}{rgb}{0.8, 0.5, 0.2}

\makeatletter
\DeclareRobustCommand\onedot{\futurelet\@let@token\@onedot}
\def\@onedot{\ifx\@let@token.\else.\null\fi\xspace}

\def\eg{\emph{e.g}\onedot}

\def\etal{\emph{et al}\onedot}
\makeatother

\definecolor{fuzzywuzzy}{rgb}{0.8, 0.4, 0.4}
\definecolor{MaceWindu}{rgb}{0.7, 0.1, 0.7}
\definecolor{indigo}{rgb}{0.29, 0.0, 0.51}
\definecolor{brown}{rgb}{0.5, 0.2, 0.0}
\definecolor{dark_green}{rgb}{0, 0.4, 0}

\newcommand{\ky}[1]{{\color{dark_green}#1}}

\definecolor{llmlight}{rgb}{0.75, 0.1, 0.75}

\renewcommand{\ky}[1]{#1}

\def \customparskip {.2em}
\renewcommand{\paragraph}[1]{\vspace{\customparskip}\noindent\textbf{#1}}

\newcommand{\bI}{\mathbf{I}}

\DeclareMathOperator*{\argmin}{arg\,min}

\newcommand{\supp}[0]{\texttt{Supplementary appendix}\xspace}

\newcommand{\flashthreed}[0]{Flash3D~\cite{szymanowicz2025flash3d}\xspace}
\newcommand{\mvsplat}[0]{MVSplat~\cite{chen2024mvsplat}\xspace}
\newcommand{\depthsplat}[0]{DepthSplat~\cite{xu2024depthsplat}\xspace}
\newcommand{\noposplat}[0]{NoPoSplat~\cite{ye2024noposplat}\xspace}
\newcommand{\pfthreeplat}[0]{PF3plat~\cite{hong2024pf3plat}\xspace}
\newcommand{\anysplat}[0]{AnySplat~\cite{jiang2025anysplat}\xspace}
\newcommand{\viewcrafter}[0]{ViewCrafter~\cite{yu2024viewcrafter}\xspace}
\newcommand{\lvsm}[0]{LVSM~\cite{jin2024lvsm}\xspace}

\newcommand{\lyra}[0]{Lyra~\cite{bahmani2025lyra}\xspace}
\newcommand{\sharpnvs}[0]{SHARP~\cite{mescheder2025sharp}\xspace}
\newcommand{\dustr}[0]{DUSt3R~\cite{dust3r}\xspace}

\newcommand{\retenk}[0]{RealEstate10K~\cite{zhou2018stereo}\xspace}
\newcommand{\dlthreedv}[0]{DL3DV-10K~\cite{dl3dv2024}\xspace}
\newcommand{\dtu}[0]{DTU~\cite{jensen2014large}\xspace}
\newcommand{\tnt}[0]{Tank \& Temples~\cite{knapitsch2017tanks}\xspace}

%% file: sections/0_abstract.tex
\begin{abstract}
While single-view novel view synthesis methods have been dramatically enhanced recently, how to quantify their performance has yet to be standardized.
In this work, we demonstrate that an important factor lies in how each method's own reference frame is aligned with the so-called ground-truth pose.
Existing methods often rely on test-time optimization of target poses for this alignment, which inherently is ill-posed due to the complexity of the optimization landscape of camera poses vs render accuracy, leading to methods being evaluated at an arbitrary local optimum.
Some recognize this limitation and thus rely on 3D reconstructions of entire scenes using training and test views to align poses, which may be infeasible.
In this work, instead, we propose to use traditional 2D-to-2D and 2D-to-3D image matching techniques to align 3D reconstructions.
We also evaluate only within co-visible regions, between different views, which is the only place where a `correct' ground truth can be defined.
Our evaluations show that with proper coordinate alignment, novel-view synthesis methods that use explicit 3D Gaussians achieve the state of the art, despite being previously reported in the literature to be outperformed by pure image-based 2D methods.
We further provide a simple baseline that directly utilizes an off-the-shelf video generator and a feed-forward geometry estimator.
Our simple baseline is already comparable to state-of-the-art methods specifically trained for novel-view synthesis, without any task-specific training.
Code is available at \url{https://github.com/ubc-vision/aligned-sv-nvs-bench}.
\end{abstract}

%% file: sections/1_introduction.tex
\section{Introduction}

Single-image novel-view synthesis is inherently challenging, as a single image provides insufficient information to fully determine the underlying 3D scene.
As a result, recent advances in the field have largely focused on injecting strong priors to hallucinate unseen geometry and appearance.
One line of work employs feed-forward models~\cite{szymanowicz2025flash3d} that reconstruct an explicit 3D representation from a single image and render novel views from it.
While effective when multiple views are available, such models often suffer from generalization issues in the single-view setting, potentially due to the difficulty of acquiring large-scale 3D training data.
Another line of work leverages diffusion-based generative models~\cite{yu2024viewcrafter,chen2025cognvs} to directly synthesize novel views conditioned on camera poses.
Benefiting from large-scale training data, these methods produce high-quality images, but frequently exhibit geometric inconsistencies across views.
More recent approaches, such as Lyra~\cite{bahmani2025lyra}, aim to combine the strengths of both paradigms by decoding camera-controlled video model representations into 3D Gaussians.
Despite the claimed advancements, it is not uncommon for a user to find that the ranking of perceived performance of methods differs from what is reported in the literature.

In this work, we show that alignment of coordinate frames is critical to properly quantifying each method's performance.
Existing protocols already recognize this: without a real-world measurement, a reconstruction is defined only up to scale~\cite{hartleybook}, and the frame of reference for the reconstruction and the ground truth may not align.
They therefore render at the ground-truth pose and rely on test-time optimization~\cite{yen2020inerf, lin2021barf, wang2021nerfmm} of the rendering camera to absorb the misalignment.
This, however, is \emph{insufficient}.
Since the photometric objective is highly non-convex, test-time optimization fails when the two coordinate frames are not nearby.
In fact, as shown in \cref{fig:intro_pose_alignment}, this happens frequently, and furthermore, as our experiments in \cref{sec:exp} show, \emph{does lead to inaccurate conclusions.}
Some methods~\cite{szymanowicz2025flash3d} thus utilize per-scene SfM reconstruction~\cite{schoenberger2016sfm} to align, but this is only possible on datasets where SfM reconstruction is available.

\input{figures/pose_est/item.tex}
%
Instead, we propose to first roughly align the two reconstructions by performing a 2D-to-2D correspondence matching~\cite{detone2018superpoint,lindenberger2023lightglue} between the rendered and the test views, then use Perspective-n-Point (PnP)~\cite{fischler1981random} to estimate the pose of the test views by relating the 3D coordinates of each pixel in the rendered view, and the corresponding 2D coordinates in the test view.
As the whole task of novel-view rendering is to generate a new view of the \emph{exactly same} scene, this method provides a robust way to align the two reconstructions.
Then, to further account for any small misalignment, we also perform test-time optimization as in the conventional protocol.
Finally, with the reconstructions aligned, we compute the standard image quality metrics
\emph{only on the co-visible regions} between the two reconstructions, computed by a geometric reprojection procedure.

The difference this brings is significant.
For example, 
Flash3D~\cite{szymanowicz2025flash3d}, which achieves 14.52 dB in terms of Peak Signal-to-Noise Ratio in RealEstate10K dataset~\cite{zhou2018stereo}, is actually 22.04 dB after proper alignment.
The latter is in line with what is reported by the authors when utilizing Structure-from-Motion to align the model's reconstruction against the entire test scene, which we remove the need for.

In addition to the revised evaluation protocol, we also provide a baseline single-image novel-view synthesis framework that distills video models into explicit 3D Gaussians.
Specifically, we use off-the-shelf video generators~\cite{wan2025wan, hacohen2024ltx} and a feed-forward geometry estimator~\cite{dust3r}.
We find that an effective strategy in utilizing these is to generate a walk-through video of an input view via video generators, then simply build a 3D Gaussian representation from these frames.
Crucially, to deal with the geometric hallucinations and drifts from video generations, we only use generated frames to initialize the locations of the 3D Gaussians, and optimize only using the single real input view.
With this simple method, we obtain results comparable to the state-of-the-art novel-view methods trained specifically for the task, \emph{without any task-specific training}.


We perform benchmarking with our new evaluation protocol on four standard datasets: RealEstate10K~\cite{zhou2018stereo}, Tanks \& Temples~\cite{knapitsch2017tanks}, DL3DV-10K~\cite{dl3dv2024}, and DTU~\cite{jensen2014large}.
Our findings are as follows:
\begin{itemize}[leftmargin=*]
    \item test-time optimization-based evaluation is not sufficient to align views;
    \item utilizing 2D-to-2D and 2D-to-3D matching can correct for these misalignments \emph{without} the need of any explicit 3D model of the entire scene;
    \item with proper alignment, 3D representation-based methods such as Flash3D~\cite{szymanowicz2025flash3d} largely outperform video based methods such as ViewCrafter~\cite{yu2024viewcrafter}, and a hybrid method that marries the best of both, Lyra~\cite{bahmani2025lyra}, performs the best;
    \item a simple baseline model that uses off-the-shelf video models~\cite{wan2025wan,hacohen2024ltx} and DUSt3R~\cite{dust3r} is comparable to the state of the art, \emph{without any task-specific training}.
\end{itemize}



%% file: figures/pose_est/item.tex
\begin{figure}
    \centering
    \includegraphics[width=\linewidth]{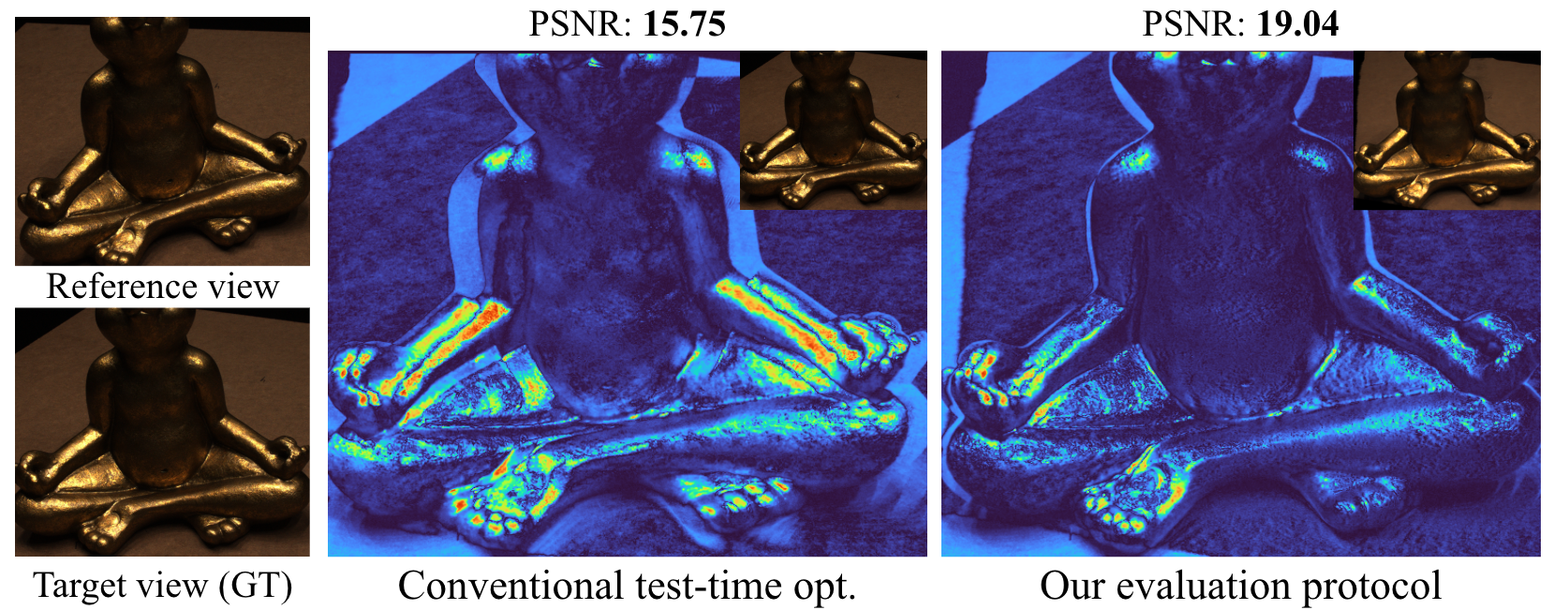}
    \vspace{-1.5em}
    \caption{
    \textbf{-- Evaluation can be misleading due to inaccurate test-time pose alignment --}
    We visualize the error map with the conventional test-time optimization and our proposed protocol (red is higher error, blue is lower).
    With the same 3D scene reconstructed from the reference view, conventional test-time optimization can leave residual misalignment between the rendered and ground-truth views, leading to large quantitative error.
    In contrast, our evaluation protocol provides a more robust alignment strategy, enabling more reliable performance measurement.
}
\vspace{-0.5em}
    \label{fig:intro_pose_alignment}
\end{figure}

%% file: sections/2_related_works.tex
\section{Related work}




We first review methods for single-view novel-view synthesis: feed-forward approaches that infer an explicit 3D scene representation, and approaches that fine-tune video diffusion models with pose conditioning. 
We then review 
evaluation protocols for single-view novel-view synthesis.


\paragraph{Single-image novel-view synthesis with 3D Gaussians.}
Driven by large-scale learned priors, recent methods have begun to address few-shot and single-view novel-view synthesis using 3D Gaussian representations.
Methods such as MV-Splat~\cite{chen2024mvsplat}, pixelSplat~\cite{charatan2024pixelsplat}, and NoPoSplat~\cite{ye2024noposplat} estimate 3D Gaussians from a small number of input views in a single forward pass.
Flash3D~\cite{szymanowicz2025flash3d} further pushes this paradigm to operate from a single input image.
Bolt3D~\cite{szymanowicz2025bolt3d} trains a diffusion model to denoise one or more views into 3D Gaussians, while Lyra~\cite{bahmani2025lyra} decodes camera-controlled video diffusion representations into 3D Gaussians given camera intrinsics estimated via MoGe~\cite{wang2025moge}.
The performance of these methods come strongly from the amount of data they were trained with, and each often requires extensive training. 
Interestingly, we show that, without task-specific training with large data, we can perform on par with these models with careful utilization of video models~\cite{wan2025wan} and feed-forward geometry estimators~\cite{dust3r}.

\paragraph{Single-image novel view synthesis in the image domain.}
Another line of work generates novel views directly in image space using diffusion or transformer-based architectures~\cite{liu2023zero, kwak2024vivid, gao2024cat3d, jin2024lvsm, rayzer}.
Zero-1-to-3~\cite{liu2023zero} trains an image-and-pose conditioned diffusion model to synthesize novel object views, which was extended in Vivid-1-to-3~\cite{kwak2024vivid} to utilize video models to improve performance.
CAT3D~\cite{gao2024cat3d} trains a multi-view diffusion model capable of generating multiple views, which can optionally be distilled into 3D Gaussians.
More recently, transformer-based architectures with camera encodings~\cite{jin2024lvsm, rayzer} directly generate novel views without explicit 3D reconstruction.
While these image-domain approaches often produce visually appealing results, they can suffer from geometric inconsistencies across views~\cite{asim25met3r}, which may lead to artifacts or blurring under significant viewpoint changes.

\paragraph{Conventional evaluation protocol.}
An unfortunate disagreement among these various methods is the way evaluation is performed.
The main disagreement comes from the fact that each 3D representation that is reconstructed has its own coordinate reference of the world (or, in the case of pure 2D ones, their implied coordinate system within the model), leading to the save `target view' meaning different physical views.
And various methods to align them has been suggested.
Some simply opt to apply test-time-optimization of camera poses with the photometric loss against the ground-truth test view, in hope that this is enough to fix the misalignment~\cite{yen2020inerf, lin2021barf, wang2021nerfmm}.
Others use either the point clouds from Structure-from-Motion (SfM)~\cite{schoenberger2016sfm} or average depth values~\cite{szymanowicz2025bolt3d}, or use RANSAC~\cite{szymanowicz2025flash3d} to scale the reconstructions.
Models that are capable of also estimating poses given multiple views utilize this to estimate the pose of the target ground-truth view and use this pose instead~\cite{zhou2025stable}.
These, however, are not applicable to the single-view case, where there is no reliable ground-truth depth or 3D reconstructions to scale against.

Closely relevant to our evaluation protocol is GS-CPR~\cite{liu2025gscpr}, where it also utilizes 2D--2D and 2D--3D correspondences with a PnP solver to perform accurate camera pose estimation for visual localization.
They also report that this standard, traditional protocol is required to achieve best performance and simply performing inference-time (test-time) optimization is not enough.
Here, we argue that this traditional process should be embedded in the evaluation protocol of novel view synthesis as well.

\paragraph{Brief review of 3D Gaussian Splatting.} 
\label{sec:3dgs}
Since many baselines, including the one that we propose, uses 3D Gaussian Splatting, we briefly review it here for completeness.
Kerbl \etal~\cite{3dgs} represent a 3D scene as a collection of anisotropic 3D Gaussians $\mathcal{G} = \{ g_i \}_{i=1}^{K}$. Each Gaussian $g_i$ is parameterized as 
$\boldsymbol{g_i} = \{ \boldsymbol{\mu_i}, \boldsymbol{\Sigma_i}, \alpha_i, \mathbf{c_i} \}$,
where $\boldsymbol{\mu_i}$ is the mean, $\boldsymbol{\Sigma_i}$ is the covariance matrix, $\alpha_i$ is the opacity, and $\mathbf{c_i}$ is the view-dependent color
modeled using spherical harmonics ~\cite{paper37}. Given a set of posed images $\{ I_n \}$, the 
3D Gaussian parameters $\mathcal{G}$ are optimized, \eg, with Markov Chain Monte-Carlo-based optimizers~\cite{3dgs-mcmc} to photometrically reconstruct the scene.

At render time, each 3D Gaussian is locally projected to the image plane as a 2D Gaussian $G_i^{2D}(\mathbf{u})$ , and its view-dependent color $\mathbf{c}_i(\mathbf{v})$ is predicted with low-order spherical harmonics. The final pixel color is obtained by sorted alpha blending of these splats,
\[
\mathbf{C}(\mathbf{u}) \;=\; \sum_{i=1}^{K} \Big[\alpha_i\, G_i^{2D}(\mathbf{u})\, \mathbf{c}_i(\mathbf{v})\Big] \prod_{j=1}^{i - 1}\!\Big(1 - \alpha_j\, G_j^{2D}(\mathbf{u})\Big),
\]
which can be performed efficiently.

%% file: sections/3_eval_protocol.tex
\section{Aligning the misaligned evaluation protocol}
\label{sec:protocol}

For a set of ground-truth test views that we wish to render and compare against, we walk through them in the order of shortest baseline, aligning each test camera pose with the reconstructed 3D representation one by one.
For example, given multiple test views for a single input view, we first render the input view, then compare against the closest test view. Then, after aligning this view according to the process described below, we proceed with the next closest test view among all aligned test views, \emph{i.e.}, we walk to the next `closest seen' view.
We detail this \supp.



In more detail, we first perform a 2D-to-2D correspondence matching~\cite{lindenberger2023lightglue} between the rendered closest seen view and the ground-truth test view.
We then form 2D-to-3D correspondences of the rendered view, which can be computed by rasterizing the 3D positions of the Gaussians onto image pixels in the same manner as the RGB values.
With these two correspondences, we then estimate the camera pose using a PnP algorithm~\cite{lepetit2009epnp}, followed by a test-time optimization of the camera pose on the co-visible regions only.
With the aligned views we compute the evaluation metrics on the co-visible mask. 
We detail each of these steps below.

\paragraph{2D-to-2D correspondences.}
\label{sec:2d2d_correspondence}
To build the 2D-to-2D correspondences between views that may differ by a wide baseline, we rely on conventional state-of-the-art feature matching methods---we specifically use LightGlue~\cite{lindenberger2023lightglue} with SuperPoint~\cite{detone2018superpoint} features.
For the application scenario at hand, where we are simply rendering novel views of the \emph{exact same} scene, these methods return reliable correspondences.
In fact, the lack of reliable correspondences would actually indicate that the rendering quality is poor.
In what comes below, we will denote these correspondences as $(\mathbf{u}_r^k, \mathbf{u}_g^k)$ where $k=1,\dots,K$ is the index of the correspondence and $\mathbf{u}_r^k$ is the 2D coordinate in the rendered view and $\mathbf{u}_g^k$ is the 2D coordinate in the ground-truth test view.


\paragraph{2D-to-3D correspondences.}
For each rendered pixel, its 3D coordinate can be obtained from the underlying scene representation, although the exact computation depends on the representation.\footnote{In cases where a 2D-to-3D correspondence cannot be formed (\eg, purely image-space methods such as \lvsm that directly generate novel views and thus expose no 3D geometry to a PnP solver), we rely on the 2D-to-2D correspondences alone and recover the relative pose from the essential matrix, resolving its scale ambiguity from the initial pose.}
Although our protocol is agnostic to the underlying 3D representation,\footnote{
For example, mesh-based models~\cite{chen2022mobilenerf} allow ray–triangle intersection, and volumetric models such as NeRF~\cite{nerf} compute expected depth along the ray, which can be back-projected. 
Monocular depth estimators may also be used when an explicit 3D model is unavailable.
}
we formulate it here based on 3DGS~\cite{3dgs}, as many recent methods adopt this representation.
Specifically, we obtain the 3D coordinate by computing the expected 3D location under the volumetric rendering weights.
Let $\mathbf{x}_i$ be the 3D center of the $i$-th Gaussian, $T_i(\mathbf{u})$ its transmittance at pixel $\mathbf{u}$ as defined in \cref{sec:3dgs}, and $\alpha_i(\mathbf{u})$ its opacity.
We write:
\begin{equation}
  \mathbf{X}(\mathbf{u}) =
  \sum_i T_i(\mathbf{u})\,\alpha_i(\mathbf{u})\,\mathbf{x}_i,
  \label{eq:raster3d}
\end{equation}
which corresponds to the visible surface location along the ray in the model's coordinate system.

\paragraph{Pose estimation via Perspective-n-Point (PnP)}
\label{sec:pnp}
Given the 2D-to-2D correspondences 
$(\mathbf{u}_r^k, \mathbf{u}_g^k)$ 
and the corresponding 3D points 
$\mathbf{X}(\mathbf{u}_r^k)$ obtained from the rendered view, 
we construct the 2D-to-3D correspondence set
\begin{equation}
  \mathcal{C} =
  \left\{
    \big(
      \mathbf{X}(\mathbf{u}_r^k),
      \mathbf{u}_g^k
    \big)
  \right\}_{k=1}^{K}.
  \label{eq:correspondences}
\end{equation}

We then estimate the pose $\mathbf{E}_0 \in \mathrm{SE}(3)$ 
of the ground-truth test view by solving a PnP problem:
\begin{equation}
  \mathbf{E}_0 =
  \mathop{\mathrm{arg\,min}}_{\mathbf{E} \in \mathrm{SE}(3)}
  \sum_{k:\,\mathbf{u}_g^k \in \Omega_{\text{c}}}
  \left\|
    \mathbf{u}_g^k -
    \pi\left(
      \mathbf{E}\,\mathbf{X}(\mathbf{u}_r^k)
    \right)
  \right\|^2,
  \label{eq:pnp_masked}
\end{equation}
where $\pi(\cdot)$ denotes the camera projection function, and 
$\Omega_{\text{c}}$ is the co-visible region between the two images, defined as a mask on the test view, which we detail below.
We solve~\eqref{eq:pnp_masked}
using a standard RANSAC-based PnP implementation,
which provides a robust initialization $\mathbf{E}_0$
for subsequent refinement via test-time optimization.

\paragraph{Determining co-visible regions.}
As not all pixels in the ground-truth test image $\bI_g$ are visible in the input image $\bI_i$, including non-visible pixels in quantifying the reconstruction quality is misleading.
We thus evaluate \emph{only} on the co-visible regions, and also perform our pose estimation with these co-visible pixels.
Importantly, we perform a \emph{conservative estimate} of the co-visible region, as missing a few actual co-visible pixels in aggregation is less harmful than including non-co-visible regions, which disproportionately disadvantages methods that do not extrapolate.

To find these pixels, we use DUSt3R~\cite{dust3r} to obtain per-frame point maps, and project pixels onto each other to determine if they are co-visible.
For typical novel-view synthesis tasks, DUSt3R provides a reliable estimate of the scene geometry, as the changes between the input image and the ground-truth test image are purely viewpoint changes.

Specifically, we perform the following steps:
\begin{itemize}[leftmargin=*]
  \item We first discard all pixels that, when projected onto the input image, are outside the image bounds.
  \item We then compute the reprojection error~\cite{vats2023gcmvsnet} by projecting each pixel of the test view to the input view with DUSt3R~\cite{dust3r} depth estimates, and projecting them back to the test view and measuring their errors, and thresholding the errors with a threshold $\tau_{\text{pixel}}$.
  \item We also compute the relative depth differences~\cite{vats2023gcmvsnet} between the depth estimated from the test view, and the depth recomputed by warping the corresponding pixel in the input view to the test view, again thresholding the errors with a threshold $\tau_{\text{depth}}$.
\end{itemize}
An example of the co-visible regions extracted by our protocol is shown in \cref{fig:coviz}.

\input{figures/coviz/item.tex}




\paragraph{Test-time refinement.}
\label{sec:tto}
After obtaining the initial pose $\mathbf{E}_0 \in \mathrm{SE}(3)$ via masked PnP,
we further refine it using conventional test-time optimization.
We parameterize incremental pose updates with Lie algebra $\mathfrak{se}(3)$
and optimize the masked photometric objective in the co-visible region:
\begin{equation}
    \min_{\boldsymbol{\xi} \in \mathfrak{se}(3)}
    \sum_{\mathbf{u} \in \Omega_{\text{c}}}
    \left\|
    \mathbf{I}_g(\mathbf{u}) -
    \mathbf{I}_r\!\left(
        \mathbf{u};
        \exp(\boldsymbol{\xi})\,\mathbf{E}_0
    \right)
    \right\|_1,
    \label{eq:tto}
\end{equation}
where $\exp(\boldsymbol{\xi}) \in \mathrm{SE}(3)$ denotes the exponential map
from the Lie algebra to the pose group,
$\mathbf{I}_g$ and $\mathbf{I}_r$ are the ground-truth and rendered images,
respectively,
and $\Omega_{\text{c}}$ is the co-visibility mask.

\paragraph{Camera trajectories for video-based methods.}
We found that directly feeding held-out test views to video-based methods~\cite{yu2024viewcrafter} causes inferior results than reported in paper.
We thus use a smooth camera trajectory to desired poses as in ViVid-1-to-3~\cite{kwak2024vivid} based on spherical-linear interpolation (Slerp).


\paragraph{Computing metrics.}
Finally, we compute the standard image quality metrics, Peak Signal-to-Noise Ratio (PSNR), Structural Similarity Index Measure (SSIM)~\cite{ssim}, and Learned Perceptual Image Patch Similarity (LPIPS)~\cite{lpips}.
All metrics are computed and averaged within the co-visible regions only.

\paragraph{Implementation details.}
We provide detailed implementation details required to reproduce our experiments in \supp.
To facilitate the adaptation of our evaluation protocol, we release our code at \url{https://github.com/ubc-vision/aligned-sv-nvs-bench}.

%% file: figures/coviz/item.tex
\begin{figure}
    \centering
    \resizebox{\linewidth}{!}{
    \setlength{\tabcolsep}{1pt}
    \begin{tabular}{ccc}
        \includegraphics[width=0.33\linewidth]{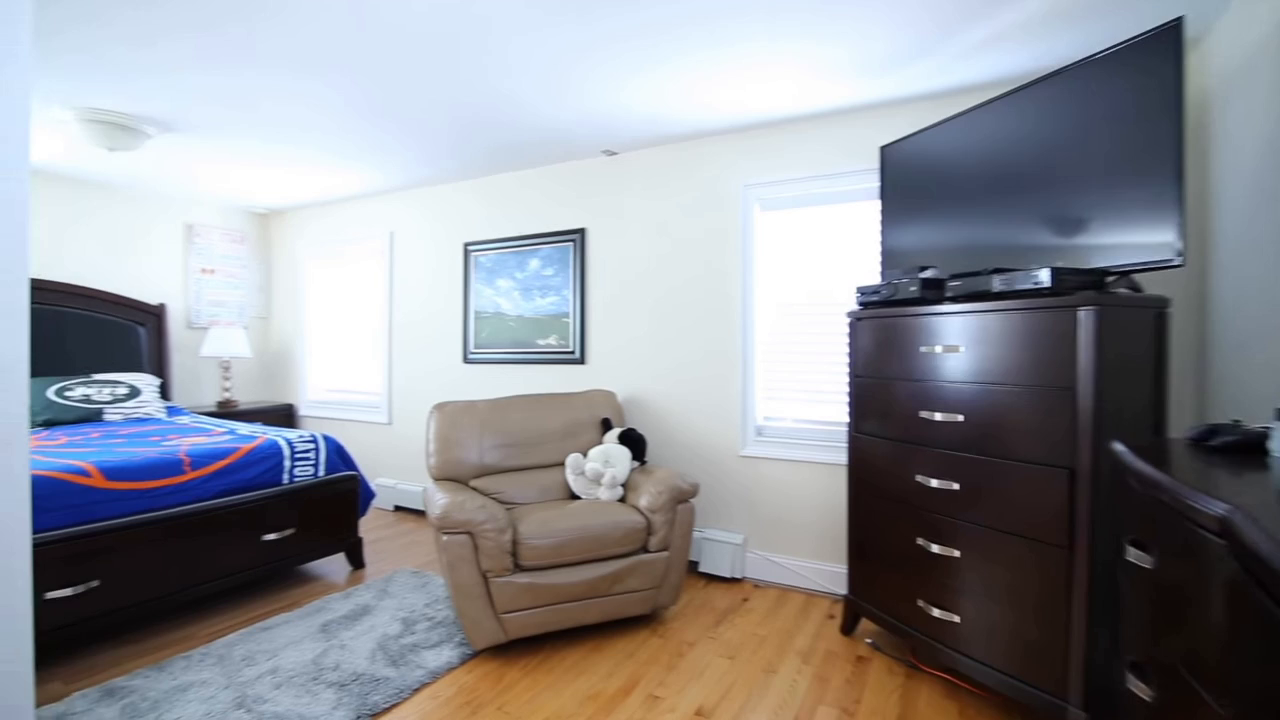} &
        \includegraphics[width=0.33\linewidth]{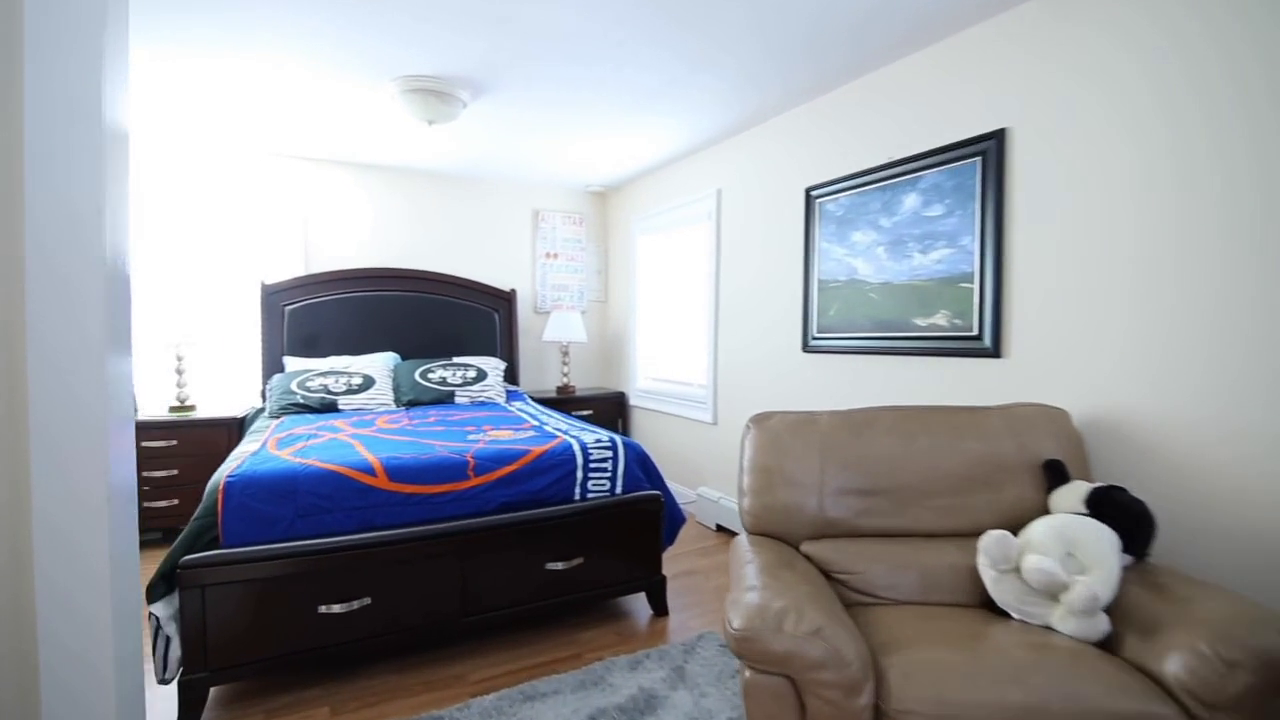} &
        \includegraphics[width=0.33\linewidth]{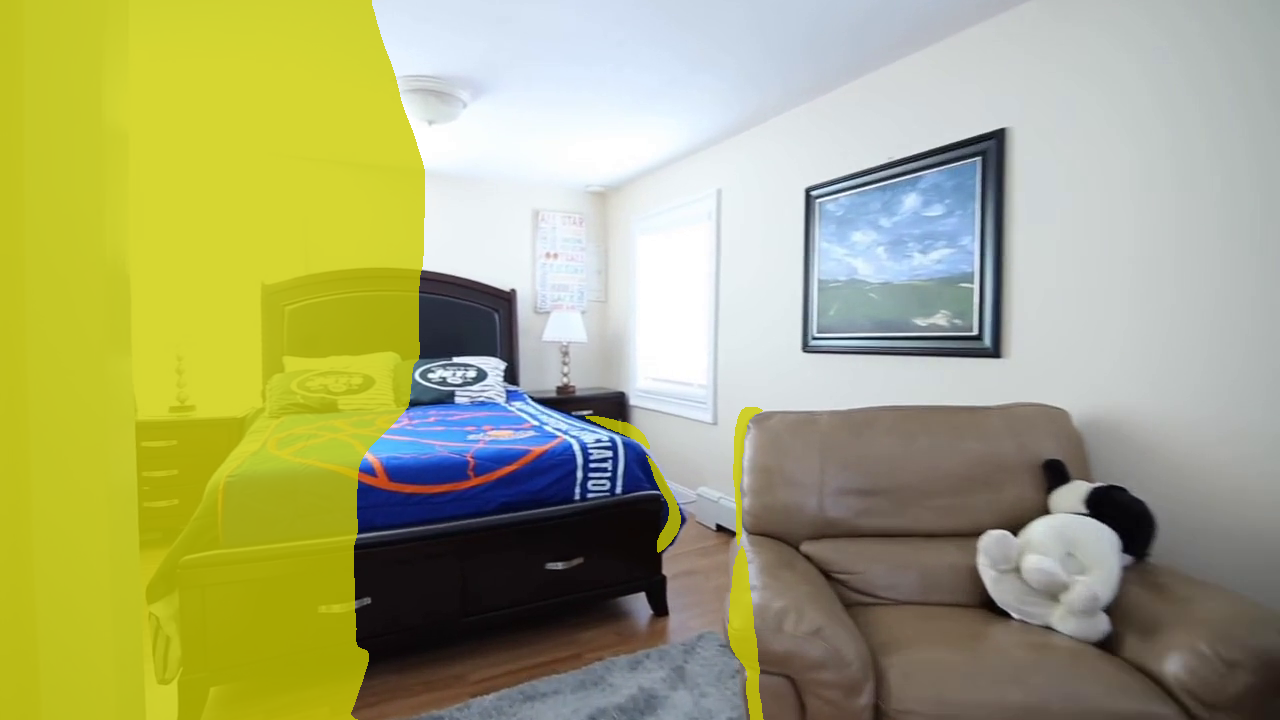} \\
        {\scriptsize (a) Input view} & {\scriptsize (b) Test view} & {\scriptsize (c) Co-visible region}
    \end{tabular}
    }
    \vspace{-0.5em}

    \caption{
        {\bf Extracting co-visible regions --}
        We show example co-visible regions extracted by our protocol.
        We extract by projecting each pixel back and forth via DUSt3R~\cite{dust3r} and checking their relative depths.
        We make a \emph{conservative} estimate of co-visible regions to not disadvantage methods that do not extrapolate pixels outside the input view.
    }
    \vspace{-.5em}
    \label{fig:coviz}
\end{figure}

%% file: sections/4_video_nvs.tex
\section{A simple baseline using video models}
\label{sec:video_nvs}

With the new evaluation protocol in place, we further propose a single-image novel-view synthesis framework to distill video models into 3D Gaussians.
We first generate a video sequence from a single input image using an off-the-shelf image-conditioned text-to-video model~\cite{wan2025wan, hacohen2024ltx} (\cref{sec:vid_gen}).
We then reconstruct a 3D Gaussian representation of the scene by initializing 3D Gaussians with generated views but optimizing only with the real input view (\cref{sec:recon}).

\subsection{Generating multiple views with video models}
\label{sec:vid_gen}

To generate novel views from a single image, we use off-the-shelf image-text-to-video models.
We use Wan~2.2~\cite{wan2025wan} for all datasets except for the DTU~\cite{jensen2014large} dataset where Wan~2.2 fails completely, and hence use LTX-Video~\cite{hacohen2024ltx}.
We synthesize a short video sequence of a camera moving in the scene, starting from the input image;
for examples, see \supp.

To generate the text prompt for the video model, we rely on an agentic vision-language model (VLM)~\cite{claude-code}.
We provide the input image to the VLM and prompt it to generate detailed scene descriptions, with additional guardrails to make the descriptions encourage video generation for static scenes.
We also add negative prompts to prevent the generation of videos with transition effects: `\textit{transition effect}' and `\textit{abrupt scene change}'.
For the exact prompt, see \supp.

\paragraph{Auto-curating video generations.}
Even with careful prompting, these models can fail to generate good videos for 3D reconstruction.
We thus automatically detect the following four types of failure modes using off-the-shelf optical flow estimators~\cite{farneback2003two,opencv_optical_flow} and regenerate if detected:
\begin{itemize}
    \item \textit{Fully static.} 
    Static videos do not provide the baseline required for 3D estimation.
    We flag a video as static when the mean optical flow magnitude across all frame pairs falls below a threshold $\tau_s$.
    \item \textit{Transition effects.}
    Video models may include transition effects such as blur or dissolve.
    We detect these via forward-backward optical-flow cycle consistency: for each pair of consecutive frames we compute dense Farneback flow in both directions and measure the per-pixel cycle-consistency error.
    To avoid false detections in textureless regions where optical flow is unreliable, we evaluate the error only over pixels whose Sobel gradient magnitude exceeds $\tau_g$.
    A video is flagged if the 95th percentile of per-frame-pair cycle errors exceeds $\tau_p$.
    \item \textit{Billboard videos.}
    Video models sometimes generate billboard videos, where the first frame is treated as a flat image.
    As billboard videos create dense optical flows that vary little, having a low coefficient of variation, we filter videos that have a coefficient of variation of dense flow lower than $\tau_f$.
\end{itemize}

We generate multiple candidate videos per scene and retain the best one that passes all checks, re-generating with a different random seed if all candidates fail.
See supplementary appendix for hyperparameters and details.

\subsection{Constructing a 3D Gaussian representation}
\label{sec:recon}

Although video diffusion models generate temporally coherent and visually plausible sequences, they are known to hallucinate geometry~\cite{voleti2024sv3d,wang2024motionctrl} even with our careful curation.
We thus opt to only use generated frames to initialize the location of 3D Gaussians, to use these generated frames as `priors', but exclude them from optimization---for optimization we only use the real input image.


\paragraph{Gaussian initialization.}
To initialize the 3D Gaussians, we use the dense point map generated by DUSt3R~\cite{dust3r}; we use all frames, including those generated by the video model.
As this results in 3D Gaussians as many as the number of pixels in each image, we uniformly
downsample to 10\% of the original number of 3D Gaussians.
For the initial scale and rotation of the Gaussians, we follow the initialization scheme in ~\cite{3dgs} 
and optimize colors up to spherical harmonics (SH) of degree 1.
Note again, by initializing this way, we loosely incorporate information generated by video models into the 3D model, and yet are tolerant to the 3D geometric errors that these video models make. 

\paragraph{Optimization.}
In order not to `drift away' due to the generated frames, we optimize only using the input view frame, which is the only real image that we can trust.
We use a standard photometric objective combining an $\ell_1$ loss and a structural similarity term~\cite{ssim}:
\begin{equation}
\mathcal{L} = (1 - \lambda_{\text{ssim}})\,\|\mathbf{I}_r - \mathbf{I}_g\|_1 + \lambda_{\text{ssim}}\,(1 - \text{SSIM}(\mathbf{I}_r, \mathbf{I}_g)),
\end{equation}
where $\mathbf{I}_r$ and $\mathbf{I}_g$ are the rendered and ground-truth images respectively, and $\lambda_{\text{ssim}} = 0.4$.
We use the Adam optimizer with the 3DGS adaptive density control strategy~\cite{3dgs}, which splits, clones, and prunes Gaussians based on view-space positional gradients.

%% file: sections/5_results.tex
\section{Experiments}
\label{sec:exp}


\vspace{-\customparskip}
\paragraph{Datasets}
We follow prior work to evaluate our model and re-evaluate baselines on single-image novel-view synthesis task across the scene datasets \retenk, \dlthreedv, \dtu, and \tnt.
For each dataset we use the test splits provided by Stable Virtual Camera~\cite{zhou2025stable} (denoted as SV), by ViewCrafter~\cite{yu2024viewcrafter} (denoted as VC), and the official test splits for the DTU dataset~\cite{jensen2014large}.
We also use the object masks from RegNeRF~\cite{niemeyer2022regnerf} for DTU following DNGaussians~\cite{li2024dngaussian} to mask out the background in addition to our co-visibility map.

Additionally, we find that the `input' view for the \dlthreedv dataset is often at one far extreme of the capture trajectory, which makes the dataset challenging, but often causes poor overlap between train and test views.
%
We thus generate our own splits and input views.
We use the camera that is most co-visible considering all other cameras,
computed as in \cref{sec:protocol}, as the input view and the rest as test views.
This causes our \dlthreedv numbers to be incompatible with what is reported in the literature.
However, our re-splitting substantially improves how much of the test views the input view covers: the average co-visible coverage rises from $45.9\%$ to $60.7\%$.

For \retenk, following Flash3D~\cite{szymanowicz2025flash3d},  we evaluate it at the resolution of 640$\times$360.
For \dtu, we use the pre-processed data from MVSplat~\cite{chen2024mvsplat}, where each image is of resolution 640$\times$512.
All other images are resized to 960$\times$540, which is the minimum resolution among the remaining images.

\paragraph{Baselines.}
We compare our simple method against feed-forward Gaussian Splatting methods~\cite{szymanowicz2025flash3d,chen2024mvsplat,xu2024depthsplat,hong2024pf3plat,jiang2025anysplat,mescheder2025sharp}, a video diffusion-based method~\cite{yu2024viewcrafter}, a pure transformer-based method~\cite{jin2024lvsm}, and a hybrid method that decodes video-diffusion representations into 3D Gaussians~\cite{bahmani2025lyra}.

We use Flash3D~\cite{szymanowicz2025flash3d} as a primary baseline for a single-view feed-forward 3D Gaussian splitting estimator.
We further include \lvsm, a transformer-based model that directly regresses the target view from the input image and target camera pose without any explicit 3D representation; as it exposes no 3D geometry for PnP, we align its poses using the 2D-to-2D essential-matrix variant of our protocol.
We also evaluate four pose-free feed-forward Gaussian Splatting methods.
\pfthreeplat regresses Gaussians together with its own camera estimates.
For \mvsplat, \depthsplat and \noposplat, we duplicate the single input view as their two context views, since they are originally developed for two-view inference---we provide them for completeness.
Note that public models for these methods operate at $256 \times 256$, thus we use this setting when evaluating these methods via center cropping on datasets.
\anysplat predicts Gaussians together with its own camera intrinsics and metric-scale geometry from unconstrained views, and we condition the method on the single input view.
We also include a recent method, \sharpnvs, which is natively monocular.

As our baseline uses a video model, we also compare against a video-diffusion-based method, ViewCrafter~\cite{yu2024viewcrafter}, which first lifts the input image to a rough 3D point cloud, renders that point cloud along the target camera trajectory, and then conditions a video diffusion model on these renders to inpaint the missing content and synthesize the novel view.
We directly provide poses and depths from our provided train-test split.
We considered CogNVS~\cite{chen2025cognvs}, which shares the point-cloud-conditioning paradigm, but decided to exclude it---CogNVS uses multiple frames to acquire scene point clouds using MegaSAM~\cite{li2025megasam}, and replacing the point cloud with a single-view estimator as in ViewCrafter~\cite{yu2024viewcrafter} did not work well.

Finally, we also include \lyra, which decodes the latent representation of a camera-controlled video diffusion model into 3D Gaussians.
\lyra is trained to use intrinsics estimated via MoGe~\cite{wang2025moge}, which we use.



\subsection{Results}

\input{figures/qualitatives/item_weird_numbers}

\vspace{-\customparskip}
\paragraph{Alignment example -- \cref{fig:weird_numbers}.}
We first show a qualitative example of aligning with test-time optimization only, and our protocol, with Flash3D~\cite{szymanowicz2025flash3d}.
As shown, with only test-time optimization, the rendered image is not aligned well with the actual target.
This is because the scene consists of various objects in various depth levels, yielding a complicated loss landscape, as the camera pose moves towards the actual aligned pose.
However, with our protocol, the initial pose from 2D-2D image matching and 2D-3D PnP is close enough to allow the poses to be well aligned.
Note how the \emph{same} reconstruction, simply because the poses are not aligned, gives wildly different PSNR values.


\input{tables/quantitative/compare}

\input{tables/quantitative/compare_square}

\input{figures/qualitatives/item_merged}

\paragraph{Quantitative results -- \cref{tab:quantitative}.}
We provide a quantitative summary in \cref{tab:quantitative}, both with test-time optimization only and our image-matching-based protocol.
A critical finding is how \flashthreed, with proper alignment, becomes one of the strongest compared methods.
For example, for the RealEstate10K (SV), Flash3D performs 14.52 dB PSNR when simply relying on test-time optimization, but this is largely due to misalignment---with our protocol, this jumps to 22.04 dB.
This clearly shows that test-time optimization alone is \emph{not} sufficient to align views.
In other words, evaluations that ignore this alignment effectively evaluate how much bias towards the test views each model's internal representation has, instead of truly evaluating the quality of the obtained 3D model.
Note also how \sharpnvs, which was the state of the art according to the commonly used protocol, is significantly worse than \lyra{}

Our suggested baseline, which uses generic video models without any training, also performs on par with \flashthreed, and outperforms the video generator-based method \viewcrafter. Among the compared methods, only \lyra{} reports a higher average.
Note that \lyra is purposely trained for novel-view synthesis, while our suggested simple baseline requires none.
Even without sophistication or training, simply filtering out generation failures and using the right combination of conventional methods delivers comparable results.

Also interesting is that \depthsplat, which seemingly performed better than \flashthreed when not properly aligned, performs worse, demonstrating that previous evaluations were measuring the alignment of internal bias within reconstructors to the datasets rather than reconstruction quality.
We emphasize once more here, though, that these methods \emph{are not designed to do} single-view novel-view synthesis and should only be taken into account as reference points.

\paragraph{Qualitative results -- \cref{fig:qualitative}.}
We show qualitative examples in \cref{fig:qualitative}. 
As shown, the qualitative results align well with what the quantitative summary reports.
\flashthreed{} and \lyra{} are the strongest baselines, with our method performing on par, sometimes performing better, and sometimes performing worse.
Note also how our conservative co-visibility mask is allowing us to evaluate only the pixels that the two images have in common.
In some cases, our co-visibility mask can be overly conservative to prevent \emph{any} non-common pixels from creeping in, which is a deliberate choice we made, as the opposite would invalidate metrics such as PSNR and SSIM, as one would be asking to hallucinate what is not provided in the input view.


\input{tables/quantitative/init_train_ablation}

\paragraph{Ablation study -- \cref{tab:init_train_ablation}.}
\label{sec:ablation}
We provide an ablation study on the usefulness of the generated frames in \cref{tab:init_train_ablation}.
As shown, best results are obtained when generated frames are used to initialize---using them for training of 3D Gaussians hurt performance as Gaussians start fitting to hallucinated geometric inaccuracies.

%% file: figures/qualitatives/item_weird_numbers.tex
\begin{figure}
    \centering
    \resizebox{\linewidth}{!}{
    \setlength{\tabcolsep}{1pt}
    \renewcommand{\arraystretch}{0.85}
    \newcommand{\wnimwidth}{0.245\linewidth}
    \begin{tabular}{cccc}
        {\scriptsize Input} &
        {\scriptsize Ground Truth Test View} &
        {\scriptsize Test-time optim. only} &
        {\scriptsize Our protocol} \\
        \includegraphics[width=\wnimwidth]{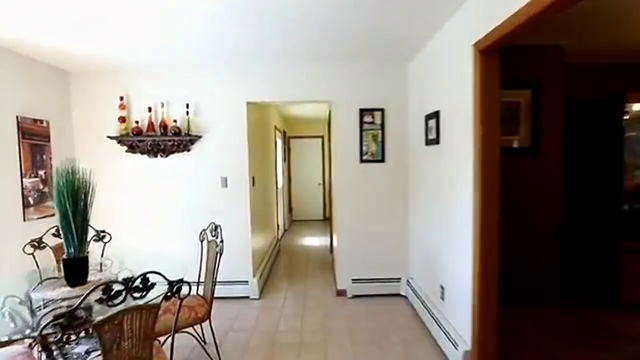} &
        \includegraphics[width=\wnimwidth]{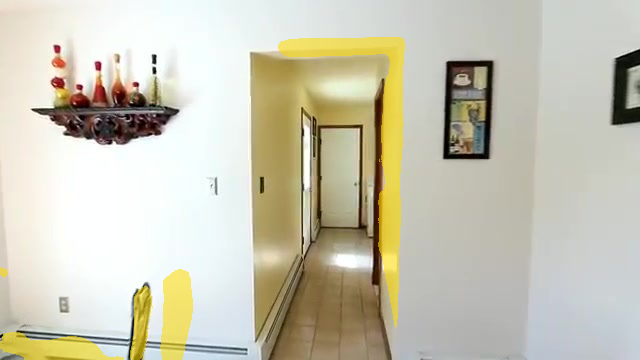} &
        \includegraphics[width=\wnimwidth]{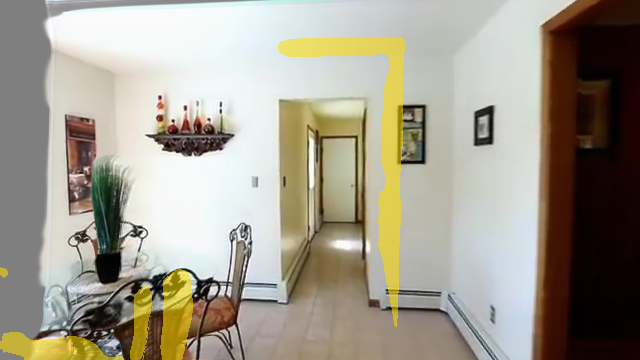} &
        \includegraphics[width=\wnimwidth]{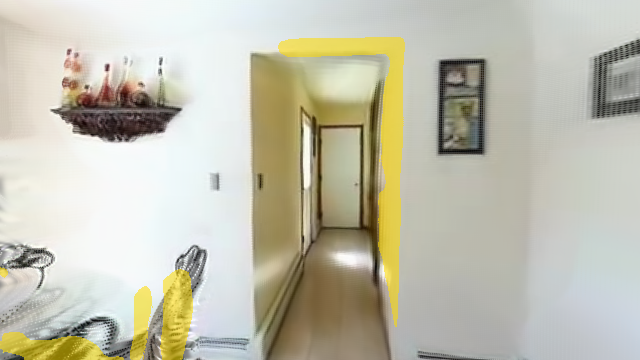} \\[1pt]
        \multicolumn{4}{c}{\scriptsize RealEstate10K, scene \texttt{01a5cc3805e94c21}} \\[2pt]
        \includegraphics[width=\wnimwidth]{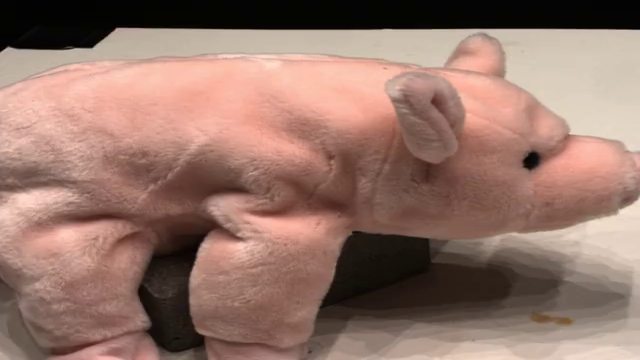} &
        \includegraphics[width=\wnimwidth]{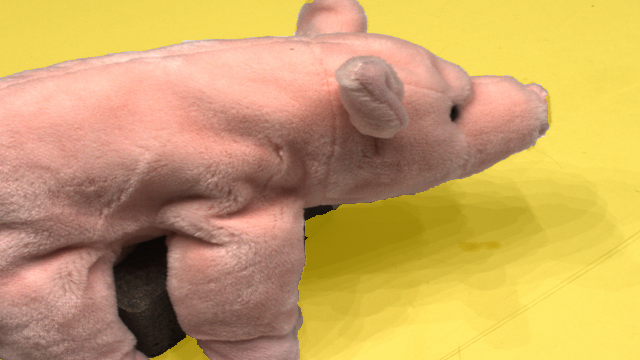} &
        \includegraphics[width=\wnimwidth]{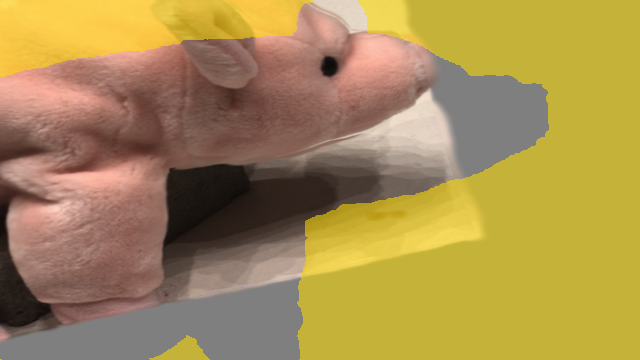} &
        \includegraphics[width=\wnimwidth]{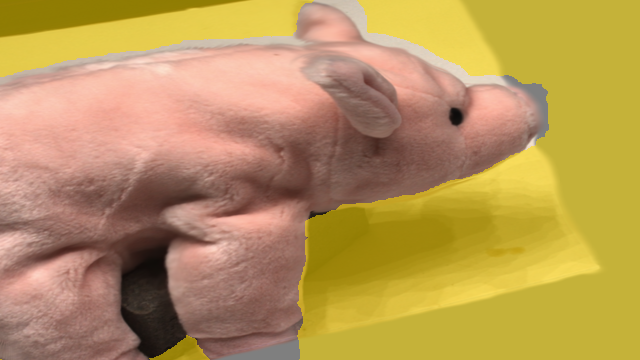} \\[1pt]
        \multicolumn{4}{c}{\scriptsize DTU, scan103} \\
    \end{tabular}
    \renewcommand{\arraystretch}{1.0}
    \setlength{\tabcolsep}{6pt}
    }
    \vspace{-.5em}
    \caption{
    {\bf Examples of misalignment leading to inaccurate evaluation --}
    We show examples with Flash3D~\cite{szymanowicz2025flash3d} for RealEstate10K (scene \texttt{01a5cc}, fr.~50) and DTU (scan103, fr.~12), with test-time optimization only, and our proposed protocol.
    As shown, optimization alone leaves a large pose offset.
    Combined with the fact that many methods do not mask via co-visibility~\cite{chen2024mvsplat, xu2024depthsplat, yu2024viewcrafter, chen2025cognvs, jin2024lvsm}, the PSNR values can be very low, up to a level where it effectively is at a level where two random images are being compared ($< 10$ dB).
    With our image-matching-based pre-alignment, on the other hand, images are better aligned, and reasonable PSNR values are extracted.
    The yellow tint marks the non-co-visible region that our protocol excludes from computing the metrics.
    }
    \vspace{-.5em}
    \label{fig:weird_numbers}
\end{figure}

%% file: tables/quantitative/compare.tex
\begin{table*}[t]
\caption{
{\bf Quantitative results -- }
We summarize the Peak Signal-to-Noise Ratio (PSNR), Structural Similarity Index Metric (SSIM)~\cite{ssim}, and Learned Perceptual Image Patch Similarity (LPIPS)~\cite{lpips} for each method on our benchmark datasets, with the mean over all datasets reported on the right.
Within each group, best results are in \textbf{bold} and second best are \underline{underlined}.
Using test-time optimization only reports wildly different results compared to when our evaluation protocol is used: Flash3D~\cite{szymanowicz2025flash3d} is among the worst performing methods with a naive evaluation but, with better alignment, performs better than DepthSplat, while SHARP~\cite{mescheder2025sharp} leads the test-time-optimization-only group on average---its Gaussians are regressed directly in the input-camera frame, so it needs little alignment to begin with.
}
\vspace{-1.5em}
\begin{center}
\label{tab:quantitative}
\setlength{\tabcolsep}{3pt}
\renewcommand{\arraystretch}{1.15}
\resizebox{\textwidth}{!}{
\begin{tabular}{cl|ccc|ccc|ccc|ccc|ccc|ccc}
\toprule
& \multirow{2}{*}{Method} &
\multicolumn{3}{c|}{\retenk (SV)} &
\multicolumn{3}{c|}{\retenk (VC)} &
\multicolumn{3}{c|}{\dlthreedv (SV)} &
\multicolumn{3}{c|}{\dtu} &
\multicolumn{3}{c|}{\tnt (VC)} &
\multicolumn{3}{c}{\textbf{Average}} \\
\cmidrule(lr){3-5}\cmidrule(lr){6-8}\cmidrule(lr){9-11}\cmidrule(lr){12-14}\cmidrule(lr){15-17}\cmidrule(lr){18-20}
& & PSNR$\uparrow$ & SSIM$\uparrow$ & LPIPS$\downarrow$ & PSNR$\uparrow$ & SSIM$\uparrow$ & LPIPS$\downarrow$ & PSNR$\uparrow$ & SSIM$\uparrow$ & LPIPS$\downarrow$ & PSNR$\uparrow$ & SSIM$\uparrow$ & LPIPS$\downarrow$ & PSNR$\uparrow$ & SSIM$\uparrow$ & LPIPS$\downarrow$ & PSNR$\uparrow$ & SSIM$\uparrow$ & LPIPS$\downarrow$ \\
\midrule
\multirow{6}{*}{\rotatebox[origin=c]{90}{\scriptsize\emph{\makecell{Test-time\\optim. only}}}}
& Ours       & 18.85 & 0.714 & 0.402 & 16.03 & 0.666 & 0.489 & 12.74 & \underline{0.341} & 0.682 & 11.83 & 0.397 & 0.647 & 14.30 & 0.454 & 0.566 & 14.75 & 0.514 & 0.557 \\
& \flashthreed & 14.52 & 0.578 & 0.504 & 14.75 & 0.630 & 0.499 & 11.71 & 0.300 & 0.683 & 11.12 & 0.402 & 0.689 & 13.75 & 0.455 & 0.574 & 13.17 & 0.473 & 0.590 \\
& \depthsplat & 16.66 & 0.612 & 0.409 & 16.45 & 0.633 & 0.421 & 12.77 & 0.300 & 0.644 & 10.67 & 0.346 & 0.623 & 13.59 & 0.441 & 0.534 & 14.03 & 0.466 & 0.526 \\
& \lyra       & 17.50 & 0.653 & 0.423 & 16.20 & 0.662 & 0.454 & 12.34 & 0.295 & 0.627 & 12.68 & \textbf{0.432} & 0.535 & \underline{15.38} & \underline{0.478} & \underline{0.482} & 14.82 & 0.504 & 0.504 \\
& \anysplat   & \underline{19.40} & \underline{0.734} & \underline{0.330} & \underline{19.03} & \underline{0.739} & \underline{0.370} & \underline{13.54} & 0.323 & \underline{0.617} & \underline{12.82} & 0.416 & \underline{0.523} & 14.81 & 0.465 & 0.508 & \underline{15.92} & \underline{0.535} & \underline{0.470} \\
& \sharpnvs   & \textbf{20.77} & \textbf{0.782} & \textbf{0.282} & \textbf{19.45} & \textbf{0.766} & \textbf{0.319} & \textbf{13.57} & \textbf{0.350} & \textbf{0.583} & \textbf{12.83} & \underline{0.431} & \textbf{0.484} & \textbf{15.55} & \textbf{0.490} & \textbf{0.427} & \textbf{16.43} & \textbf{0.564} & \textbf{0.419} \\
\midrule
\multirow{6}{*}{\rotatebox[origin=c]{90}{\scriptsize\emph{\makecell{Our eval.\\protocol}}}}
& Ours       & \textbf{22.26} & \textbf{0.808} & 0.270 & \textbf{21.30} & \textbf{0.807} & 0.280 & 14.13 & \underline{0.396} & 0.532 & \underline{17.47} & 0.461 & 0.456 & 16.80 & 0.554 & 0.405 & \underline{18.39} & \underline{0.605} & 0.389 \\
& \flashthreed & \underline{22.04} & \underline{0.788} & 0.278 & 20.77 & 0.768 & 0.329 & 13.51 & 0.371 & 0.582 & 17.06 & \underline{0.470} & 0.469 & \underline{17.68} & \underline{0.574} & 0.421 & 18.21 & 0.594 & 0.416 \\
& \depthsplat & 17.82 & 0.649 & 0.350 & 17.64 & 0.668 & 0.357 & 13.20 & 0.333 & 0.561 & 13.51 & 0.352 & 0.547 & 14.39 & 0.468 & 0.455 & 15.31 & 0.494 & 0.454 \\
& \lyra       & 21.66 & 0.786 & \underline{0.262} & \underline{21.02} & \underline{0.798} & \underline{0.272} & \textbf{14.68} & \textbf{0.427} & \underline{0.495} & \textbf{17.98} & \textbf{0.479} & 0.464 & \textbf{19.27} & \textbf{0.632} & \textbf{0.320} & \textbf{18.92} & \textbf{0.624} & \underline{0.362} \\
& \anysplat   & 20.72 & 0.747 & 0.275 & 19.80 & 0.731 & 0.325 & 14.24 & 0.347 & 0.529 & 17.18 & 0.412 & \underline{0.446} & 16.73 & 0.497 & 0.398 & 17.74 & 0.547 & 0.394 \\
& \sharpnvs   & 21.29 & 0.772 & \textbf{0.246} & 20.03 & 0.770 & \textbf{0.270} & \underline{14.41} & 0.393 & \textbf{0.493} & 16.16 & 0.419 & \textbf{0.420} & 16.84 & 0.519 & \underline{0.334} & 17.74 & 0.575 & \textbf{0.353} \\
\midrule
\rotatebox[origin=c]{90}{\scriptsize\emph{GT pose}}
& \viewcrafter & 14.34 & 0.610 & 0.405 & 16.70 & 0.687 & 0.324 & 13.32 & 0.356 & 0.503 & 13.85 & 0.357 & 0.495 & 16.42 & 0.529 & 0.361 & 14.93 & 0.508 & 0.418 \\
\bottomrule
\end{tabular}
}
\end{center}
\vspace{-.5em}
\end{table*}

%% file: tables/quantitative/compare_square.tex
\begin{table*}[t]
\caption{
{\bf Quantitative results (256$\times$256 square evaluation) -- }
PSNR, SSIM~\cite{ssim}, and LPIPS~\cite{lpips} for the methods that support the square (256$\times$256 centre-crop) setting, under both test-time-optimization-only and our evaluation protocol, with the mean over all datasets reported on the right.
\flashthreed\ and \viewcrafter\ are omitted as they cannot produce a comparable 256$\times$256 cell.
\pfthreeplat is evaluated with its RealEstate10K checkpoint on every dataset, as are \mvsplat and \noposplat.
Within each group, best results are in \textbf{bold} and second best are \underline{underlined}.
}
\vspace{-1.5em}
\begin{center}
\label{tab:quantitative_square}
\setlength{\tabcolsep}{3pt}
\renewcommand{\arraystretch}{1.15}
\resizebox{\textwidth}{!}{
\begin{tabular}{cl|ccc|ccc|ccc|ccc|ccc|ccc}
\toprule
& \multirow{2}{*}{Method} &
\multicolumn{3}{c|}{\retenk (SV)} &
\multicolumn{3}{c|}{\retenk (VC)} &
\multicolumn{3}{c|}{\dlthreedv (SV)} &
\multicolumn{3}{c|}{\dtu} &
\multicolumn{3}{c|}{\tnt (VC)} &
\multicolumn{3}{c}{\textbf{Average}} \\
\cmidrule(lr){3-5}\cmidrule(lr){6-8}\cmidrule(lr){9-11}\cmidrule(lr){12-14}\cmidrule(lr){15-17}\cmidrule(lr){18-20}
& & PSNR$\uparrow$ & SSIM$\uparrow$ & LPIPS$\downarrow$ & PSNR$\uparrow$ & SSIM$\uparrow$ & LPIPS$\downarrow$ & PSNR$\uparrow$ & SSIM$\uparrow$ & LPIPS$\downarrow$ & PSNR$\uparrow$ & SSIM$\uparrow$ & LPIPS$\downarrow$ & PSNR$\uparrow$ & SSIM$\uparrow$ & LPIPS$\downarrow$ & PSNR$\uparrow$ & SSIM$\uparrow$ & LPIPS$\downarrow$ \\
\midrule
\multirow{6}{*}{\rotatebox[origin=c]{90}{\scriptsize\emph{\makecell{Test-time\\optim. only}}}}
& Ours       & \textbf{20.73} & \textbf{0.744} & 0.315 & 16.08 & 0.538 & 0.462 & \underline{13.04} & \underline{0.260} & 0.668 & 11.82 & 0.340 & 0.652 & \underline{14.09} & 0.364 & 0.536 & \underline{15.15} & 0.449 & 0.527 \\
& \lvsm      & \underline{19.76} & \underline{0.690} & \textbf{0.283} & \textbf{19.54} & \textbf{0.672} & \textbf{0.270} & \textbf{13.37} & \textbf{0.266} & \underline{0.632} & \underline{12.58} & \textbf{0.372} & 0.602 & \textbf{16.16} & \textbf{0.443} & \textbf{0.440} & \textbf{16.28} & \textbf{0.489} & \underline{0.445} \\
& \mvsplat   & 15.51 & 0.589 & 0.401 & 14.54 & 0.535 & 0.404 & 11.91 & 0.227 & 0.657 & 10.34 & 0.275 & 0.643 & 12.41 & 0.351 & 0.563 & 12.94 & 0.395 & 0.534 \\
& \depthsplat & 17.47 & 0.656 & 0.360 & \underline{16.10} & \underline{0.602} & 0.378 & 12.39 & 0.247 & 0.648 & 12.01 & \underline{0.369} & \underline{0.539} & 13.49 & \underline{0.399} & 0.517 & 14.29 & \underline{0.455} & 0.488 \\
& \noposplat & 13.42 & 0.466 & 0.538 & 12.29 & 0.404 & 0.584 & 11.31 & 0.249 & 0.740 & 9.41 & 0.327 & 0.801 & 11.51 & 0.295 & 0.623 & 11.59 & 0.348 & 0.657 \\
& \pfthreeplat & 18.25 & 0.660 & \underline{0.313} & 15.91 & 0.589 & \underline{0.333} & 10.59 & 0.171 & \textbf{0.605} & \textbf{12.82} & 0.314 & \textbf{0.524} & 12.95 & 0.395 & \underline{0.446} & 14.10 & 0.426 & \textbf{0.444} \\
\midrule
\multirow{6}{*}{\rotatebox[origin=c]{90}{\scriptsize\emph{\makecell{Our eval.\\protocol}}}}
& Ours       & \textbf{22.29} & \textbf{0.785} & \textbf{0.250} & \textbf{21.40} & \textbf{0.760} & \underline{0.213} & \textbf{14.73} & \underline{0.369} & \underline{0.490} & \textbf{17.49} & \textbf{0.479} & \textbf{0.379} & \textbf{17.40} & \textbf{0.543} & \textbf{0.324} & \textbf{18.66} & \textbf{0.587} & \textbf{0.331} \\
& \lvsm      & \underline{20.17} & 0.706 & \underline{0.254} & \underline{20.72} & \underline{0.715} & \textbf{0.201} & \underline{13.74} & 0.314 & 0.535 & 15.45 & 0.413 & 0.486 & \underline{17.36} & \underline{0.520} & \underline{0.340} & \underline{17.49} & \underline{0.534} & \underline{0.363} \\
& \mvsplat   & 16.49 & 0.667 & 0.307 & 15.71 & 0.636 & 0.290 & 13.12 & \textbf{0.406} & \textbf{0.466} & 13.23 & 0.349 & 0.484 & 13.75 & 0.507 & 0.384 & 14.46 & 0.513 & 0.386 \\
& \depthsplat & 17.95 & 0.692 & 0.307 & 16.81 & 0.646 & 0.309 & 12.53 & 0.316 & 0.539 & 14.64 & 0.396 & 0.448 & 14.15 & 0.463 & 0.421 & 15.22 & 0.503 & 0.405 \\
& \noposplat & 18.31 & 0.661 & 0.325 & 17.25 & 0.618 & 0.327 & 12.81 & 0.322 & 0.538 & 13.16 & 0.387 & 0.486 & 14.02 & 0.427 & 0.439 & 15.11 & 0.483 & 0.423 \\
& \pfthreeplat & 18.41 & \underline{0.713} & 0.276 & 15.83 & 0.642 & 0.294 & 11.50 & 0.251 & 0.519 & \underline{16.51} & \underline{0.429} & \underline{0.427} & 13.61 & 0.482 & 0.370 & 15.17 & 0.503 & 0.377 \\
\bottomrule
\end{tabular}
}
\end{center}
\vspace{-0.5em}
\end{table*}

%% file: figures/qualitatives/item_merged.tex
\begin{figure*}
    \centering
    \setlength{\tabcolsep}{1pt}
    \renewcommand{\arraystretch}{0.85}
    \newcommand{\rowh}{0.069\linewidth}
    \newcommand{\qim}[1]{\includegraphics[height=\rowh]{figures/qualitatives/#1}}
    \begin{tabular}{@{}c@{\hspace{6pt}}ccccc@{\hspace{6pt}}ccc@{}}
        & \multicolumn{5}{c}{\scriptsize\textbf{Native resolution}} &
        \multicolumn{3}{c}{\scriptsize\textbf{$256 \times 256$ square}} \\[1pt]
        {\tiny GT} &
        {\tiny Flash3D} & {\tiny ViewCrafter} & {\tiny Lyra} & {\tiny SHARP} & {\tiny \bf Ours} &
        {\tiny LVSM} & {\tiny PF3plat} & {\tiny \bf Ours} \\[2pt]
        \multicolumn{9}{@{}l}{\scriptsize\textbf{RealEstate10K}~\cite{zhou2018stereo} \textbf{(SV)} \enspace \textcolor{gray}{\tiny scene: 000c3ab189999a83, test id 31}} \\[2pt]
        \qim{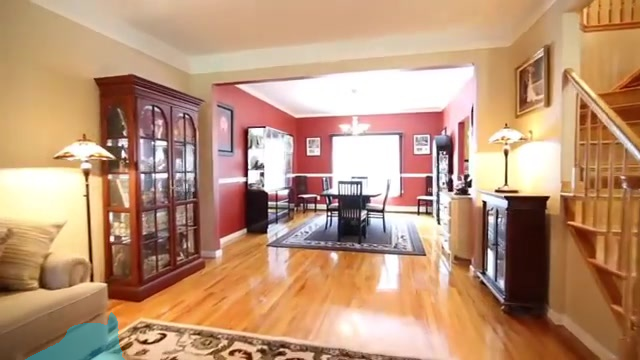} &
        \qim{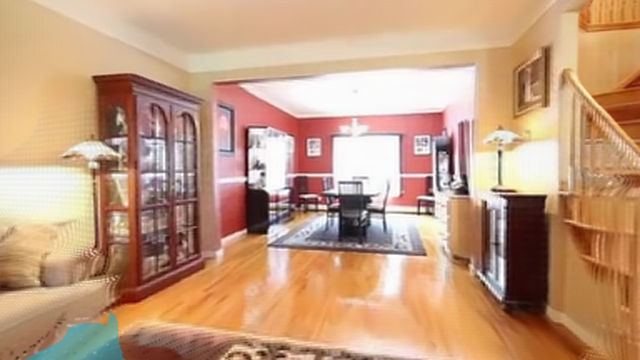} &
        \qim{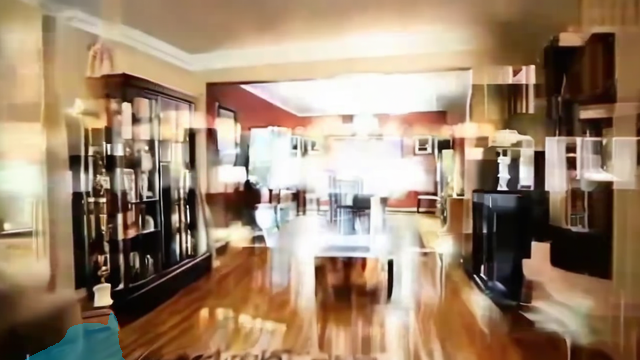} &
        \qim{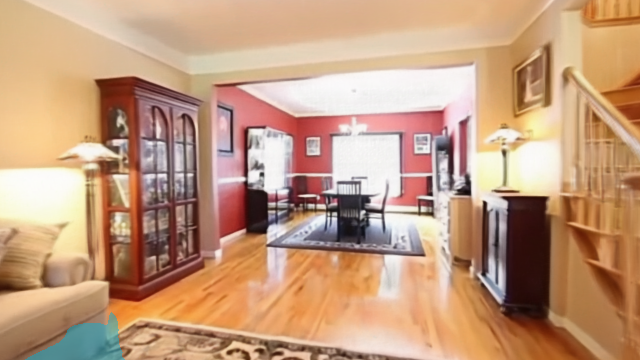} &
        \qim{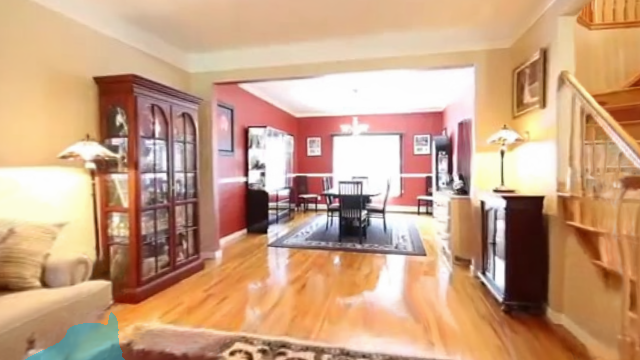} &
        \qim{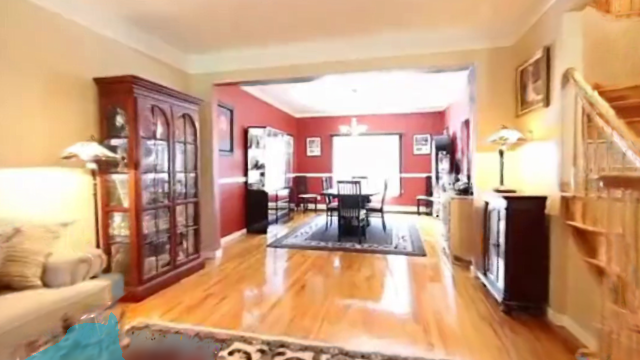} &
        \qim{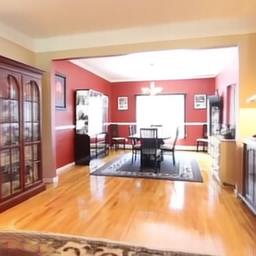} &
        \qim{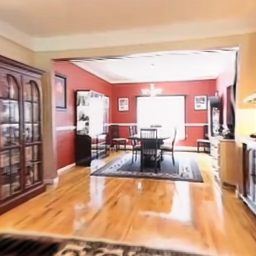} &
        \qim{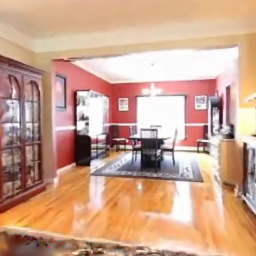} \\[2pt]
        \multicolumn{9}{@{}l}{\scriptsize\textbf{RealEstate10K}~\cite{zhou2018stereo} \textbf{(VC)} \enspace \textcolor{gray}{\tiny scene: 1, test id 4}} \\[2pt]
        \qim{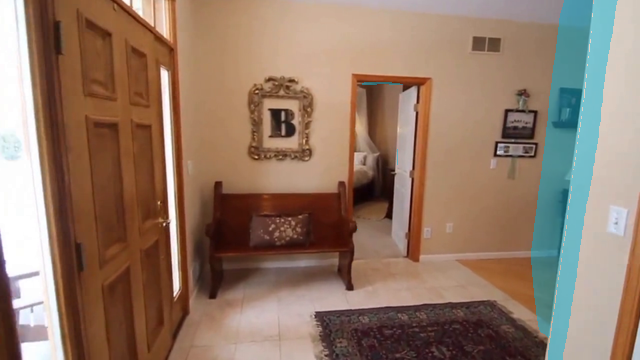} &
        \qim{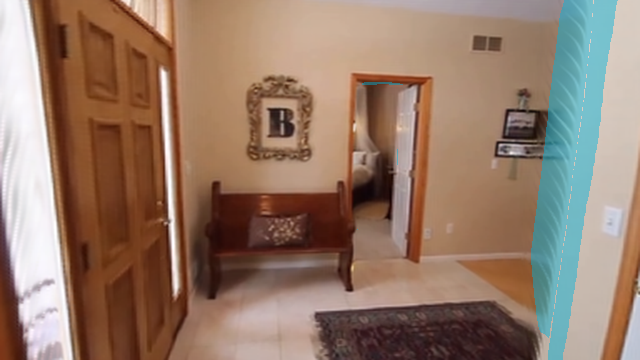} &
        \qim{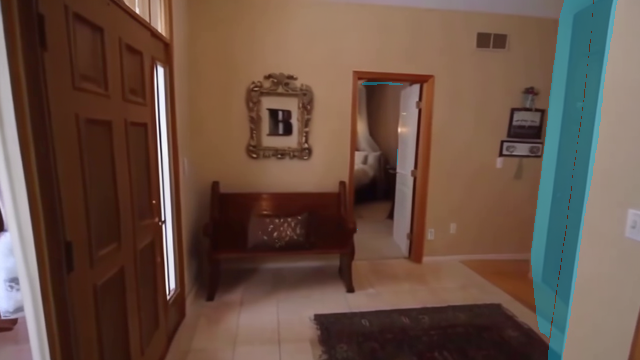} &
        \qim{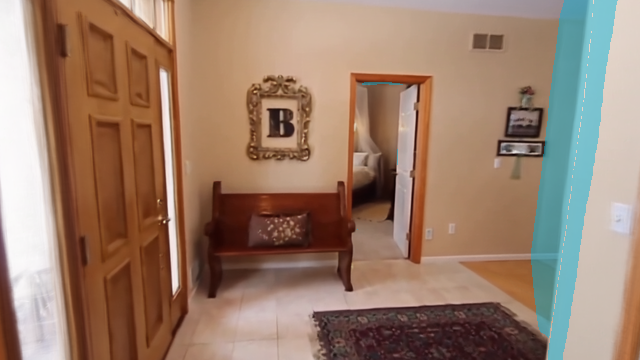} &
        \qim{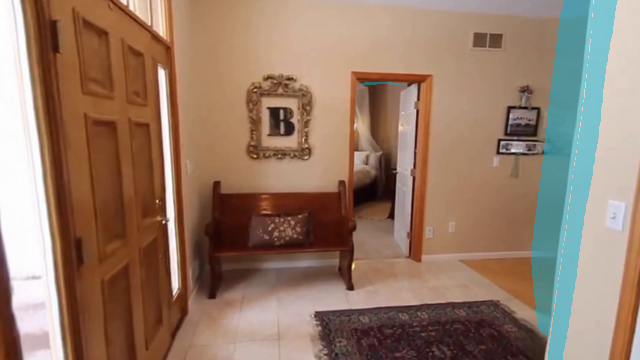} &
        \qim{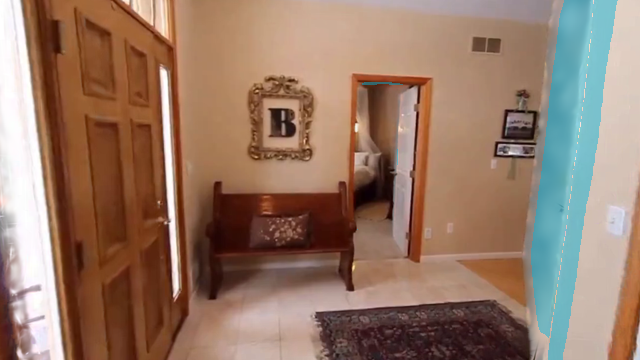} &
        \qim{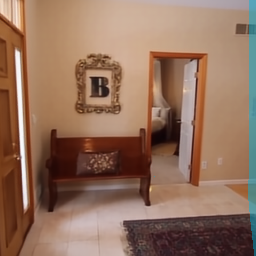} &
        \qim{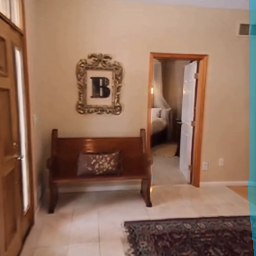} &
        \qim{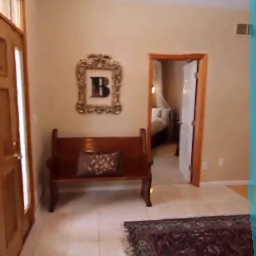} \\[2pt]
        \multicolumn{9}{@{}l}{\scriptsize\textbf{DL3DV-10K}~\cite{dl3dv2024} \textbf{(SV)} \enspace \textcolor{gray}{\tiny scene: 165f5af8, test id 17}} \\[2pt]
        \qim{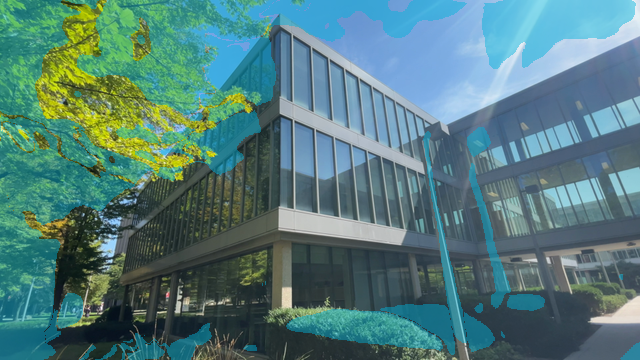} &
        \qim{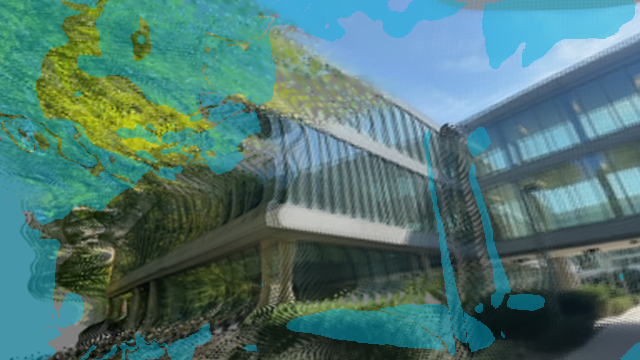} &
        \qim{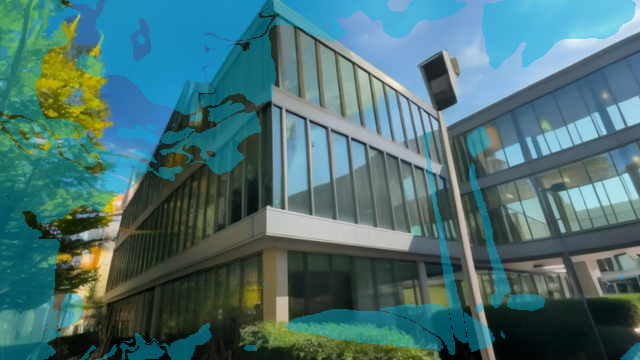} &
        \qim{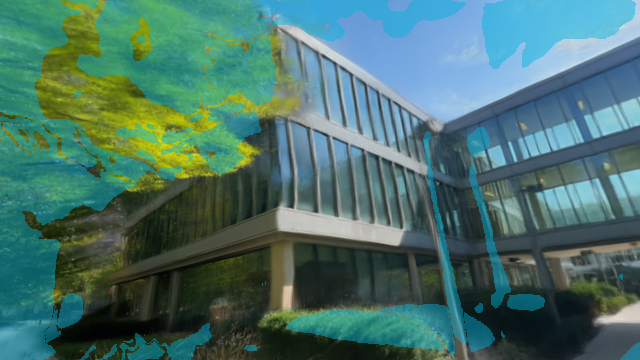} &
        \qim{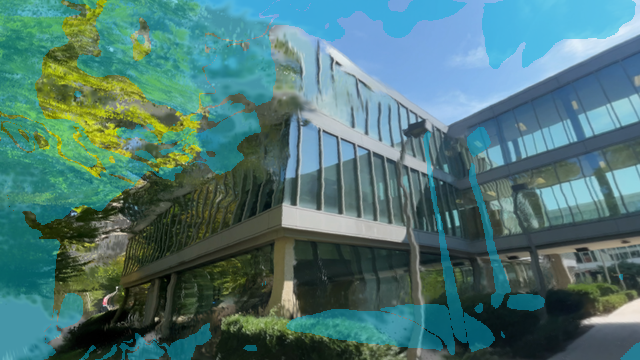} &
        \qim{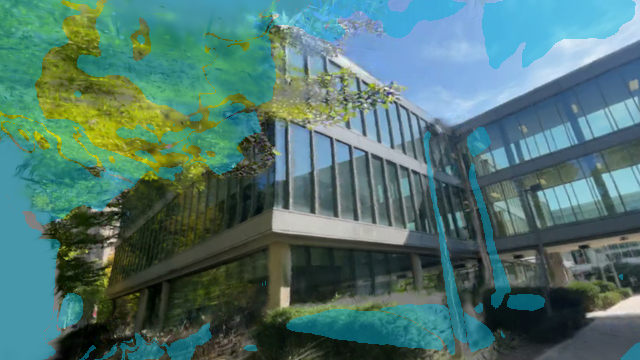} &
        \qim{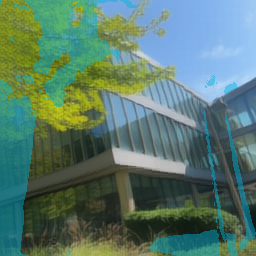} &
        \qim{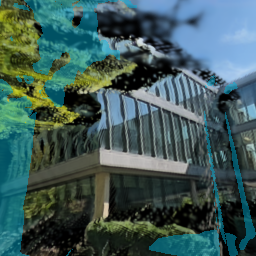} &
        \qim{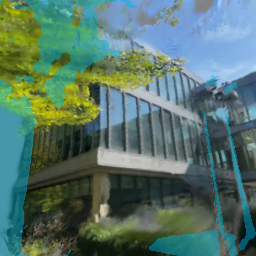} \\[2pt]
        \multicolumn{9}{@{}l}{\scriptsize\textbf{DTU}~\cite{jensen2014large} \enspace \textcolor{gray}{\tiny scene: scan45, test id 10}} \\[2pt]
        \qim{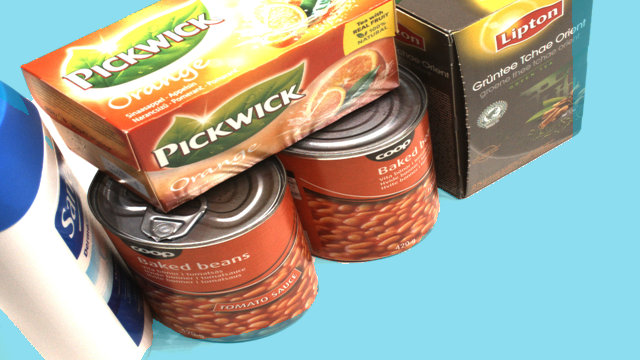} &
        \qim{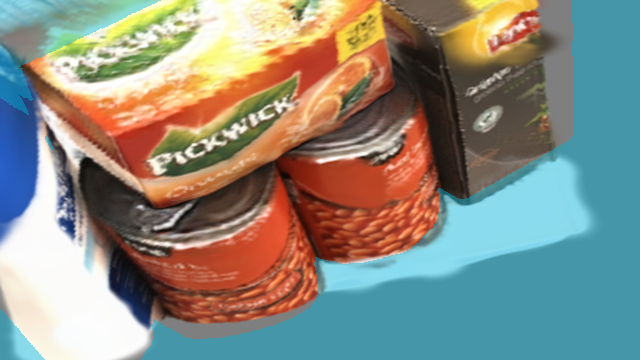} &
        \qim{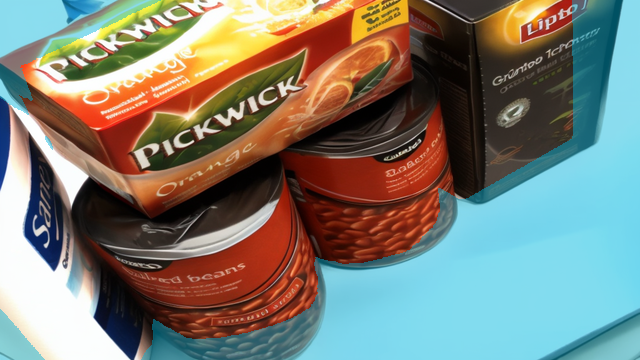} &
        \qim{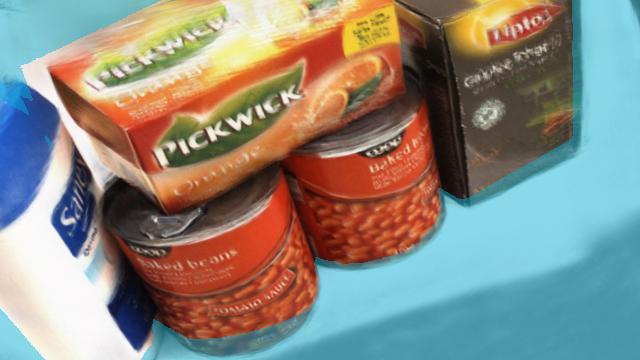} &
        \qim{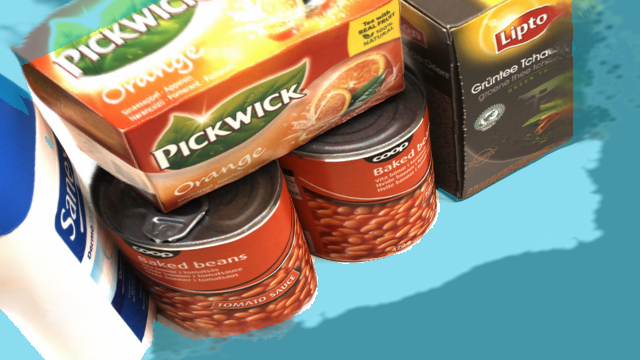} &
        \qim{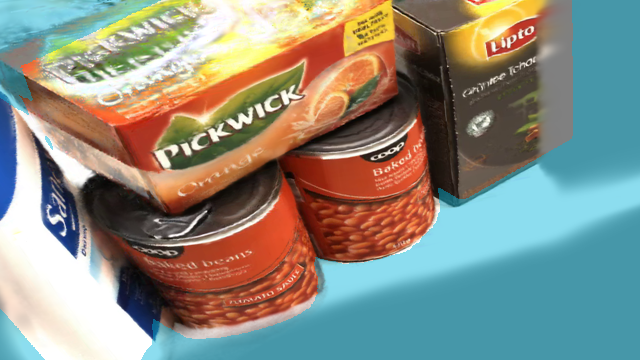} &
        \qim{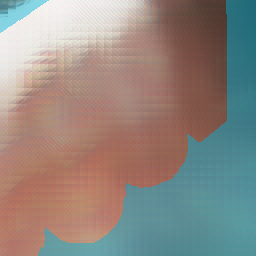} &
        \qim{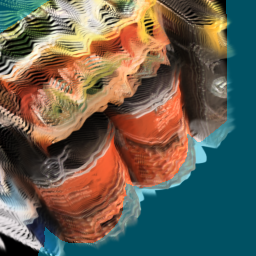} &
        \qim{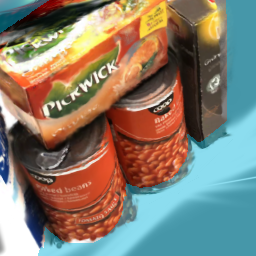} \\[2pt]
        \multicolumn{9}{@{}l}{\scriptsize\textbf{Tanks \& Temples}~\cite{knapitsch2017tanks} \textbf{(VC)} \enspace \textcolor{gray}{\tiny scene: Auditorium\_1, test id 5}} \\[2pt]
        \qim{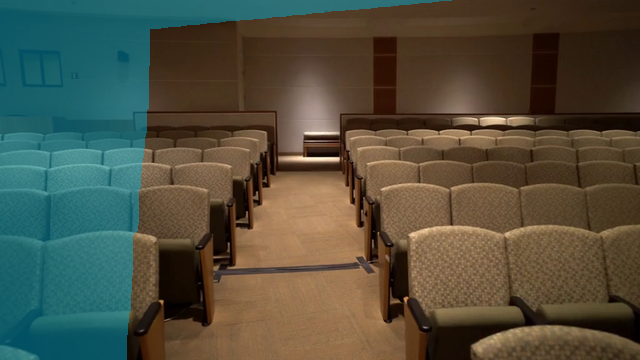} &
        \qim{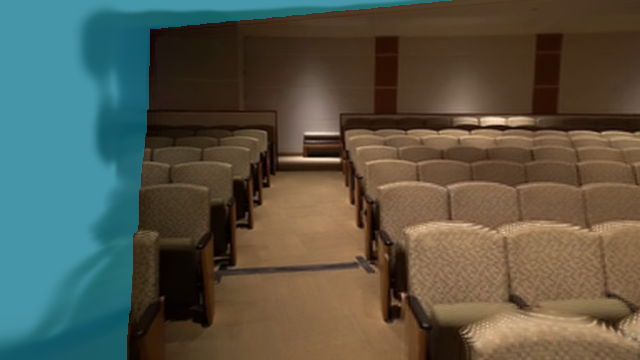} &
        \qim{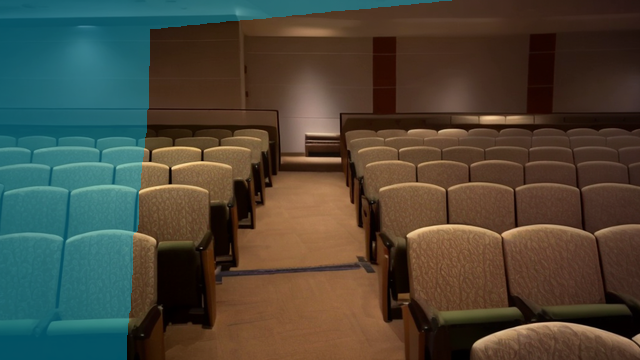} &
        \qim{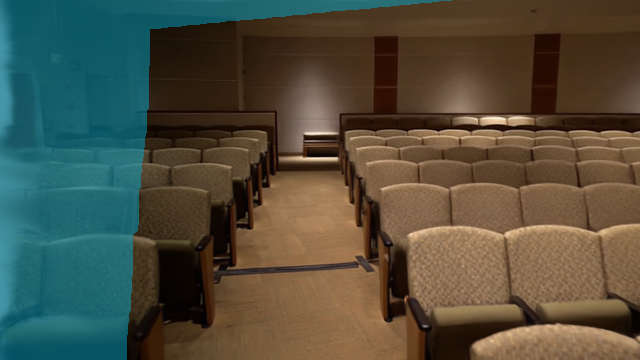} &
        \qim{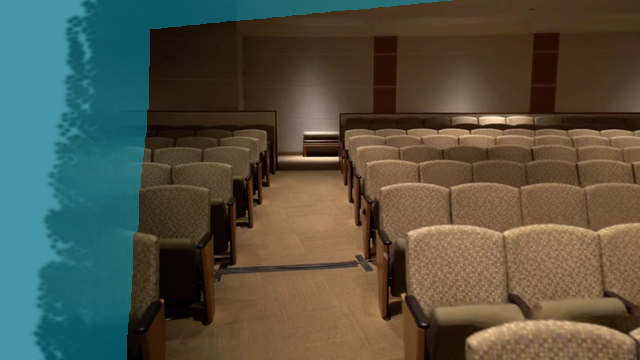} &
        \qim{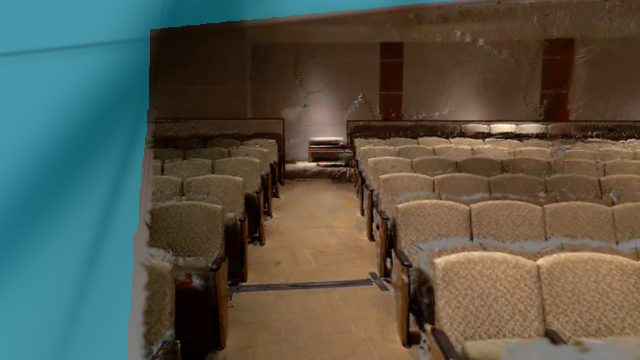} &
        \qim{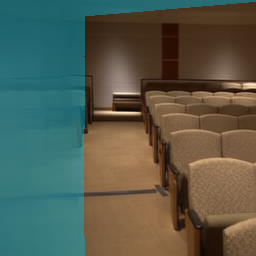} &
        \qim{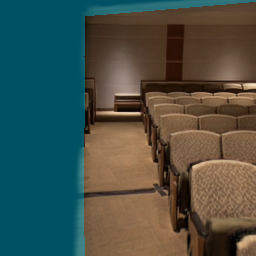} &
        \qim{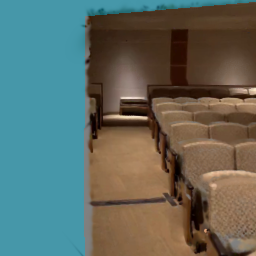} \\
    \end{tabular}
    \renewcommand{\arraystretch}{1.0}
    \setlength{\tabcolsep}{6pt}
    \vspace{-0.5em}

    \caption{
        {\bf Qualitative results -- }
        We show qualitative examples for each dataset under both evaluation resolutions, with the ground-truth target view (co-visibility overlay in blue) on the left.
        \emph{Middle:} native-resolution methods, \flashthreed, \viewcrafter, Lyra~\cite{bahmani2025lyra}, \sharpnvs, and our own baseline (\cref{tab:quantitative}).
        \emph{Right:} methods that support the $256 \times 256$ centre-crop setting, \lvsm, \pfthreeplat, and our baseline (\cref{tab:quantitative_square}).
        Both halves show the same scene and held-out target view for each dataset, listed above each row.
        All methods are shown with our conservative co-visibility overlay.
        With our enhanced alignment protocol, the actual qualitative performances of \flashthreed, \viewcrafter, and our baseline align well with the quantifications in \cref{tab:quantitative}.
        Our suggested baseline, while simple, performs on par with \flashthreed.
    }
    \label{fig:qualitative}
\end{figure*}

%% file: tables/quantitative/init_train_ablation.tex
\begin{table}
\centering
\caption{%
{\bf Ablation on using generated views -- }
We ablate when to use generated views for training our 3D Gaussians.
Including generated views for initialization only and training Gaussians only with the real input view gives best results.
}
\vspace{-.5em}
\label{tab:init_train_ablation}
\setlength{\tabcolsep}{4pt}
\renewcommand{\arraystretch}{1.1}
\resizebox{\columnwidth}{!}{
\begin{tabular}{llcccccc}
\toprule
Init & Train & RE10K~\cite{zhou2018stereo} & RE10K~\cite{zhou2018stereo}-VC & T\&T~\cite{knapitsch2017tanks} & {\small \dlthreedv} & \dtu & Avg \\
\midrule
input only    & input only  & 21.34 & 20.11 & 16.32 & 13.61 & 16.56 & 17.59 \\
all views & input only  & \textbf{22.26} & \textbf{21.30} & 16.80 & \textbf{14.13} & \textbf{17.47} & \textbf{18.39} \\
input only    & all views & 19.89 & 20.17 & 17.24 & 13.46 & 16.24 & 17.40 \\
all views & all views & 19.92 & 20.34 & \textbf{17.40} & 13.60 & 16.41 & 17.53 \\
\bottomrule
\end{tabular}
}
\vspace{-.5em}
\end{table}

%% file: sections/6_conclusion.tex
\section{Conclusion}
\label{sec:conclusion}

We have introduced a robust evaluation protocol for single-view novel-view synthesis that properly aligns 3D reconstructions using traditional image matching and PnP, and evaluates only on co-visible regions. 
Importantly, because we rely on traditional image matching techniques, our protocol does not require any 3D models of the scene to be present.
Our evaluations reveal that relying only on test-time optimization can be largely misleading, particularly due to the misalignments that cannot be corrected by it.
We also provide a simple baseline that uses off-the-shelf video generators and feed-forward geometry estimators to perform single-view novel-view synthesis, which is comparable to the state of the art without any task-specific training.
We hope our evaluation protocol will help standardize future comparisons in this field.

\paragraph{Limitations and future work.}
While our findings could also be applicable to other novel-view synthesis tasks other than single-view novel-view synthesis, we have not been able to explore them yet.
For example, few-shot novel-view synthesis and dynamic novel-view synthesis may also suffer from similar issues, as they also often rely on test-time optimization in their evaluation. 
Introducing our evaluation protocol to those tasks could be an interesting future work.


%% file: sections/X_supplementary.tex
\clearpage

\appendix
\maketitlesupplementary


\section{Implementation details for the benchmark}

\paragraph{`Closest seen' views.}
To apply 2D-to-2D followed by 2D-to-3D matching, we must render a view that shares enough in common.
We thus walk through the test views, starting with the shortest baselines from the input view.
Concretely, 
Let $\mathbf{o}_c \in \mathbb{R}^3$ denote the ground-truth camera center of view $c$, and let $\mathbf{P}_c \in \mathrm{SE}(3)$ denote its pose in the model's estimated world.
A view is \emph{seen} once it has been assigned an estimated-world pose; we collect these in the set $\mathcal{S}$, initialized with the single input view $v_0$, whose estimated-world pose is known by construction.
The render view for a target $t$ is the seen view nearest to $t$ in the ground-truth world, $s^\star = \argmin_{s \in \mathcal{S}} \lVert \mathbf{o}_t - \mathbf{o}_s \rVert_2$, and after $t$ is localized via PnP and test-time optimization its estimated-world pose is registered into $\mathcal{S}$ for subsequent targets.
Targets are processed in a rolling nearest-camera (Prim-style) order over $\{\mathbf{o}_c\}$.
\Cref{alg:closest-seen} summarizes the procedure.

\paragraph{2D-to-2D.}
For SuperPoint~\cite{detone2018superpoint} features we use 2048 maximum number of keypoints with default settings.
For LightGlue~\cite{lindenberger2023lightglue}, we also use the default setting of 9 attention layers with adaptive depth and width pruning enabled (depth confidence = 0.95, width confidence = 0.99).

\paragraph{2D-to-3D.}
For PnP, we use OpenCV's iterative Levenberg--Marquardt solver inside a RANSAC~\cite{fischler1981random} loop, with an inlier reprojection threshold of 4 pixels, at most $10^3$ iterations,\footnote{
More iterations did not significantly improve results.
}
and a confidence of $0.99$; we require at least 6 inliers for the estimate to be accepted, and the 2D-to-2D correspondences fed to PnP are restricted to the co-visible region $\Omega_{\text{c}}$.

\paragraph{Final test-time optimization}
For the final test-time optimization, we parameterize the pose update in $\mathfrak{se}(3)$ and use the Adam~\cite{kingma2014adam} optimizer with a learning rate of $5 \times 10^{-3}$, a weight decay of $10^{-4}$, and a cosine-annealing schedule decaying to $10^{-4}$ over 100 iterations, keeping the iterate with the lowest masked photometric loss (\cref{eq:tto}), which is computed over $\Omega_{\text{c}}$ only.

\paragraph{Co-visible mask computation.}
For DUSt3R~\cite{dust3r}, we use the ViT-L encoder with the DPT head at the 512-pixel input resolution, and register all frames of a scene with its global point-cloud alignment (300 iterations, cosine learning-rate schedule from $10^{-2}$, minimum confidence threshold of 5).
We compute the co-visibility masks with the cycle-consistency check of \cref{sec:protocol}: each test-view pixel is projected into the input view using the DUSt3R depths and reprojected back, and is kept only if the round-trip pixel displacement is below $\tau_{\text{pixel}} = 100$ pixels and the relative depth difference between the reprojected and the original test-view depth, $|d_{\text{reproj}} - d| / d$, is below $\tau_{\text{depth}} = 0.4$.
We evaluate both checks at the DUSt3R processing resolution, and upsample the resulting mask to the evaluation resolution with nearest-neighbor interpolation.

\begin{algorithm}[t]
\caption{`Closest seen' view selection.}
\label{alg:closest-seen}
\begin{algorithmic}[1]
\footnotesize
\Require input view $v_0$ with known estimated-world pose $\mathbf{P}_{v_0}$; target views $\mathcal{T}$; ground-truth camera centers $\{\mathbf{o}_c\}$
\State $\mathcal{S} \gets \{ v_0 \}$ \Comment{seen set: views with an estimated-world pose}
\State order $\mathcal{T}$ by rolling nearest-camera traversal over $\{\mathbf{o}_c\}$
\For{each target $t \in \mathcal{T}$ (in order)}
  \State $s^\star \gets \argmin_{s \in \mathcal{S}} \lVert \mathbf{o}_t - \mathbf{o}_s \rVert_2$ \Comment{closest \emph{seen} in world coord.}
  \State $I_{s^\star} \gets \mathrm{Render}(\mathbf{P}_{s^\star})$ \Comment{render from $s^\star$'s estimated-world pose}
  \State $\{(\mathbf{u}_r^k, \mathbf{u}_g^k)\} \gets \mathrm{Match}(I_{s^\star}, I_t^{\mathrm{gt}})$ \Comment{2D-to-2D with $\mathbf{u}_g^k \in \Omega_c$}
  \State $\mathbf{E}_0 \gets \mathrm{PnP}(\{(\mathbf{X}(\mathbf{u}_r^k), \mathbf{u}_g^k)\})$ \Comment{est. target pose in world coord.}
  \State $\mathbf{P}_t \gets \mathrm{TTO}(\mathbf{E}_0; I_t^{\mathrm{gt}}, \Omega_c)$ \Comment{photometric refinement}
  \State $\mathcal{S} \gets \mathcal{S} \cup \{ t \}$ \Comment{register $t$ as a render view for later targets}
\EndFor
\end{algorithmic}
\end{algorithm}

\section{Implementation details for our simple baseline}
\label{supp:vid_gen}

\input{figures/vid_gen/item.tex}

\paragraph{Prompt for generating videos.}
We use an agentic vision language model ~\cite{claude-code} to generate image captions that serve as text prompts for the video diffusion model.
We instruct it with the following prompt:

\textit{
Manually generate a caption in several sentences and at most 55 words for every image in \$DIRECTORY. Put a caption.txt to each image's folder. I need to use this caption as the prompt input for an image-to-video diffusion model. For the caption, begin with a description of camera motion, which is slow, lateral translation with explicit parallax cues, and should not be purely tilt, pan, rotation, or zoom. Then describe the scene content that the input image shows. End the caption with a statement that the scene is static and humanless, and for scenes containing plants, you should put that the plants stay completely still without wind, and for scenes containing vehicle or machinery, you should add that they are parked and stationary.
}

We also use the negative prompt of the classifier-free guidance to discourage videos not suitable for our task, which is "\textit{bad quality, inconsistent motion, blurry, jittery, distorted, transition effect, abrupt scene change, dolly zoom}.''.

\paragraph{Automated curation of video generations.}
We use the following thresholds for our three-stage filtering pipeline. 
For static detection, we set $\tau_s = 0.4$ on the mean Farneback optical flow~\cite{farneback2003two} magnitude computed at $256 \times 256$ resolution. 
For transition detection, we compute the cycle-consistency error only over pixels with image gradient magnitude above $\tau_g = 10$ to avoid wrong detection in textureless regions where optical flow is unreliable, and flag a video as having a transition effect if the 95th percentile of per-frame-pair cycle errors exceeds $\tau_p = 0.25$. 
For billboard detection, we set $\tau_f = 0.3$ on the coefficient of variation of flow magnitude computed at $512 \times 512$ resolution. 
We generate nine candidate videos per scene and retain the first that passes all checks.

\Cref{fig:vid_gen} shows examples of generated videos given an input view.

\paragraph{3D Gaussian training.}
We initialize the 3D Gaussians from the uniformly downsampled raw DUSt3R~\cite{dust3r} pointmaps of all frames (\cref{sec:recon}).
Each Gaussian is initialized at its point location with the point's color as the degree-0 spherical-harmonics coefficient, an isotropic scale equal to the mean distance to its three nearest neighbors, a random orientation, and an opacity of $0.1$; the degree-1 coefficients are zero-initialized and enabled after $1{,}000$ steps.
We optimize for $7{,}000$ steps with Adam~\cite{kingma2014adam}, with no depth or opacity/scale regularization.
Per-parameter learning rates follow standard 3DGS~\cite{3dgs} practice: $1.6\times10^{-4}$ for positions, $5\times10^{-3}$ for scales, $5\times10^{-2}$ for opacities, $10^{-3}$ for orientations, and $2.5\times10^{-3}$ / $1.25\times10^{-4}$ for degree-0 / degree-1 spherical harmonics.
For densification we use the standard clone-and-split strategy~\cite{3dgs} with its default $30{,}000$-step schedule (refinement every $100$ steps between steps $500$ and $15{,}000$, opacity reset every $3{,}000$ steps) linearly rescaled by $7{,}000/30{,}000$ to our step budget, giving refinement every $23$ steps between steps $116$ and $3{,}500$ and opacity reset every $700$ steps; Gaussians with opacity below $5\times10^{-3}$ are pruned.

\section{Effect of generated views -- Qualitative examples}

\input{figures/init_train_ablation/item}

\Cref{fig:init_train_ablation_qual} provides qualitative examples for the ablation of \cref{tab:init_train_ablation} (\cref{sec:ablation}), covering the full factorial over whether the generated views are used for initialization and for training.
The qualitative behavior mirrors the quantitative one: supervising the 3D Gaussians on the generated views smears fine geometry and hallucinates floaters in disoccluded regions, as the diffusion-generated frames are not multi-view consistent, whereas using all generated views for initialization still sharpens geometry over initializing from the input view alone.
The best configuration is thus initializing with the generated views while training with the input view only, matching \cref{tab:init_train_ablation}.

\section{Input view selection for \dlthreedv}
\label{sec:supp_dl3dv}

For \dlthreedv\ we replace the benchmark input view with the camera of maximum co-visibility over the train$\cup$test union.
We build the full $N \times N$ geometric co-visibility matrix $C$ on the cached DUSt3R~\cite{dust3r} depths, where $C_{ij}$ is the fraction of frame $i$'s pixels that are cycle-consistent when reprojected into frame $j$, using the same consistency checker and thresholds as the canonical mask build (pixel threshold $100$, relative depth threshold $0.4$).
Because co-visibility is asymmetric, the relevant per-camera score is the column mean
\begin{equation}
    \mathrm{as\_tgt}(c) = \frac{1}{N}\sum_i C_{ic},
\end{equation}
that is, the average over all other frames $i$ of the \emph{fraction of frame $i$'s pixels} that remains co-visible when observed in $c$---a pixel-level quantity, not a count of cameras.
We take the input view to be $\arg\max_c \mathrm{as\_tgt}(c)$, with every remaining camera becoming a test view.
Note this is the column mean rather than the row mean $\mathrm{as\_src}(c) = \frac{1}{N}\sum_j C_{cj}$, which measures how much of $c$ is visible elsewhere; the two differ because $C$ is not symmetric.
This criterion directly improves coverage: the mean $\mathrm{as\_tgt}$ co-visibility of the test views in the input view rises from $45.9\%$ for the original benchmark input view to $60.7\%$ for ours, averaged over the ten scenes, with an improvement on every scene.
All other datasets use their benchmark-provided input view.

\section{Implementation details for \viewcrafter}

\ky{%
While a public implementation exists, a `proper' application of \viewcrafter is non-trivial as there are details that can impact the outcome substantially.
We provide these below.
}%

\input{figures/viewcrafter_stages/item}

\paragraph{Point-cloud conditioning.}
\viewcrafter rasterizes a single-image point cloud from the target pose and hands the result to a video-diffusion model that synthesizes the final view (\cref{fig:viewcrafter_stages}).
\ky{This conditioning point cloud is created} by unprojecting the single input view with the canonical DUSt3R~\cite{dust3r} depth and camera intrinsics of our protocol---the same geometry every method in our benchmark is evaluated against---so its point cloud shares the protocol's metric scale and is rendered from the protocol's target poses.
The raster is geometrically and photometrically faithful wherever the single input view provides coverage, so the synthesized geometry is well-anchored near the input; as the camera rotates away, the diffusion model must hallucinate the uncovered regions and can drift in color and structure.

\paragraph{Target pose.}
\viewcrafter\ conditions on a point cloud rasterized at the \emph{target} pose, so the pose used to render that conditioning frame directly determines whether the output can align with the reference view at which the metric is computed.
Natively, \viewcrafter\ obtains this pose by running \dustr\ jointly over the reference \emph{and} target images and reading the target extrinsics from the globally aligned reconstruction; the point cloud and the conditioning render therefore share \dustr's coordinate frame by construction.
This has two consequences for evaluation. 
First, the native protocol consumes the target image to recover its pose---information a pose-conditioned method is not otherwise given---so it is not directly comparable to methods supplied with benchmark poses.
Second, the estimate inherits \dustr's failure modes: as the baseline grows and inter-view overlap drops, the global alignment degrades, the rasterized content is misplaced, and the conditioning frame no longer corresponds to the pose at which the metric is evaluated.
We therefore render \viewcrafter\ directly from the canonical DUSt3R-estimated target poses supplied by our protocol---the same poses every method is evaluated against---which removes target-pose estimation error as a confound, so the numbers reported in \cref{tab:quantitative} reflect view-synthesis quality rather than pose-estimation accuracy.
Because the conditioning point cloud is built in the same canonical DUSt3R frame (see above), the protocol poses apply directly, with no additional alignment, pose estimation, or test-time refinement.
This is also the more favorable setting for \viewcrafter: the canonical pose yields a conditioning raster correctly aligned with the target view at every baseline, whereas PnP$+$TTO would additionally penalize it for pose drift.

%% file: figures/vid_gen/item.tex
\begin{figure*}[!t]
    \centering
    \setlength{\tabcolsep}{2pt}
    \renewcommand{\arraystretch}{0.85}
    \resizebox{\linewidth}{!}{
    \begin{tabular}{cccccc}
        \includegraphics[width=0.167\linewidth]{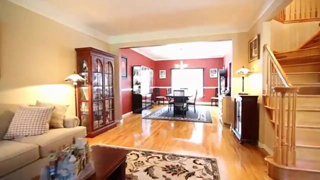} &
        \includegraphics[width=0.167\linewidth]{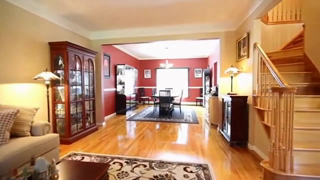} &
        \includegraphics[width=0.167\linewidth]{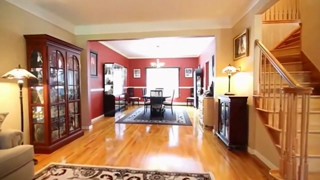} &
        \includegraphics[width=0.167\linewidth]{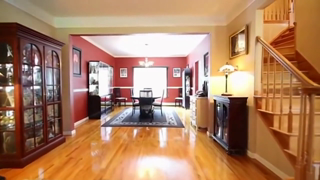} &
        \includegraphics[width=0.167\linewidth]{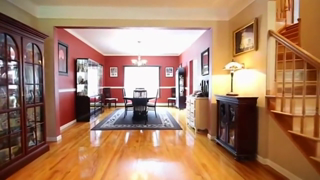} &
        \includegraphics[width=0.167\linewidth]{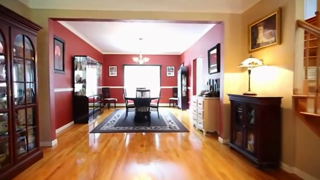} \\
        {\scriptsize Input view} & {\scriptsize Gen. view 1} & {\scriptsize Gen. view 2} & {\scriptsize Gen. view 3} & {\scriptsize Gen. view 4} & {\scriptsize Gen. view 5}
    \end{tabular}
    }
    \renewcommand{\arraystretch}{1.0}
    \setlength{\tabcolsep}{6pt}
    \vspace{-0.5em}

    \caption{
        {\bf Example views generated by video models --}
        \ky{%
        We show example views generated by video models.
        }%
    }
    \label{fig:vid_gen}
\end{figure*}

%% file: figures/init_train_ablation/item.tex
\begin{figure*}
\centering
\includegraphics[width=\linewidth]{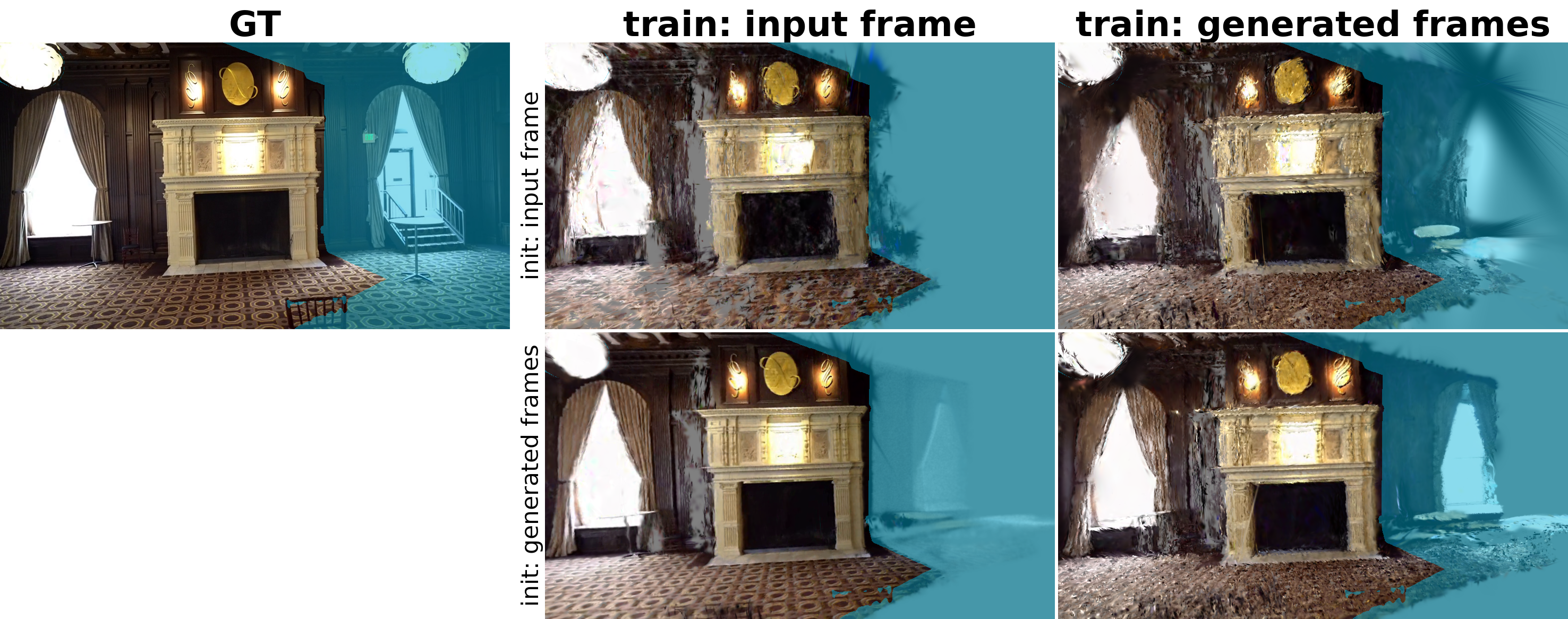}
\vspace{2pt}
\includegraphics[width=\linewidth]{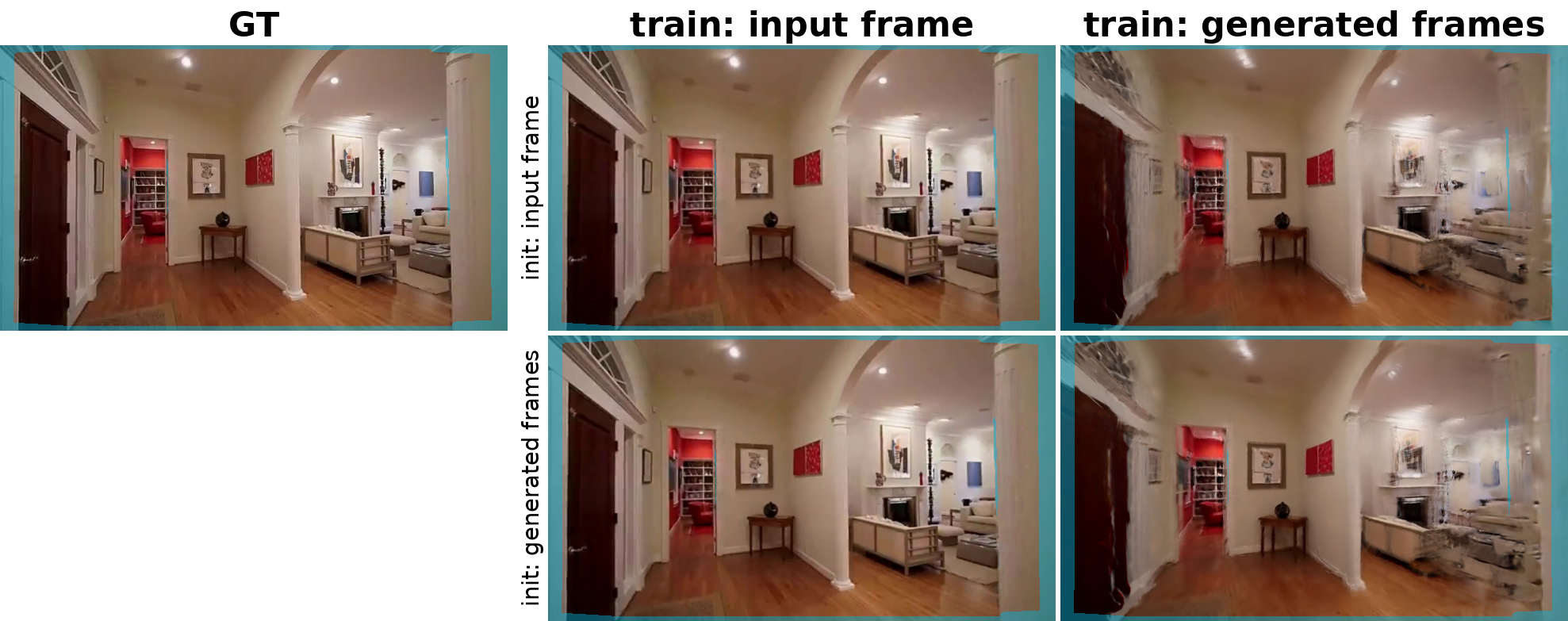}
\caption{%
{\bf Qualitative effect of generated views -- }
For each scene we show the ground truth and the full init$\times$train factorial of \cref{tab:init_train_ablation}: rows vary the initialization (input view only vs.\ all generated views) and columns the training supervision.
Training on generated views smears fine geometry (top scene) and hallucinates floaters in disocclusions (bottom scene); using them only for initialization stays sharpest.
The teal overlay marks pixels outside the co-visible region, excluded from the metric.
}
\label{fig:init_train_ablation_qual}
\end{figure*}

%% file: figures/viewcrafter_stages/item.tex
\begin{figure}[t]
    \centering
    \newcommand{\vsw}{0.315\linewidth}
    \setlength{\tabcolsep}{1pt}
    \renewcommand{\arraystretch}{0.85}
    \begin{tabular}{ccc}
        \includegraphics[width=\vsw]{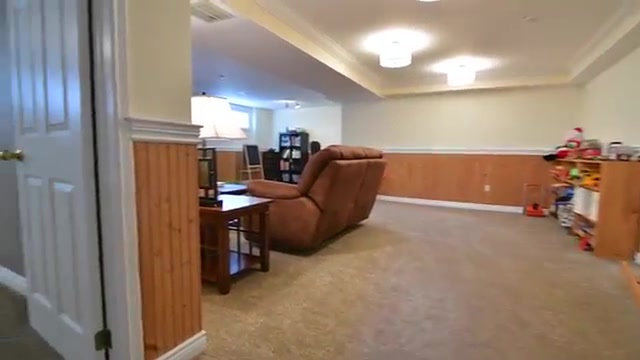} &
        \includegraphics[width=\vsw]{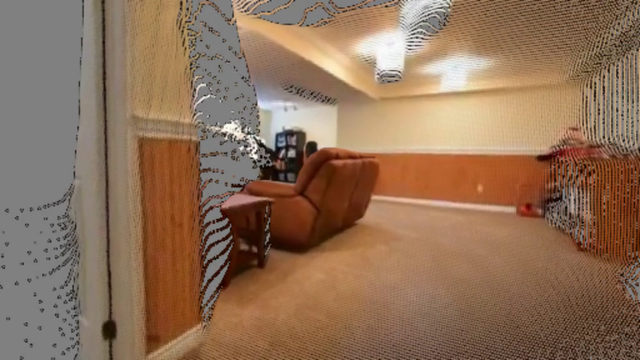} &
        \includegraphics[width=\vsw]{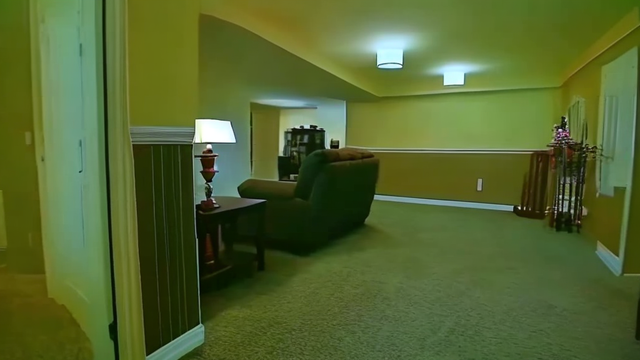} \\[1pt]
        \includegraphics[width=\vsw]{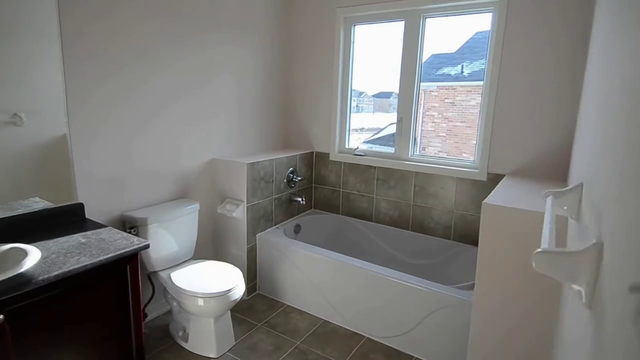} &
        \includegraphics[width=\vsw]{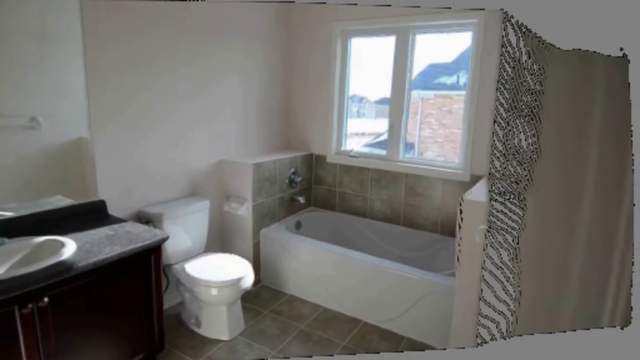} &
        \includegraphics[width=\vsw]{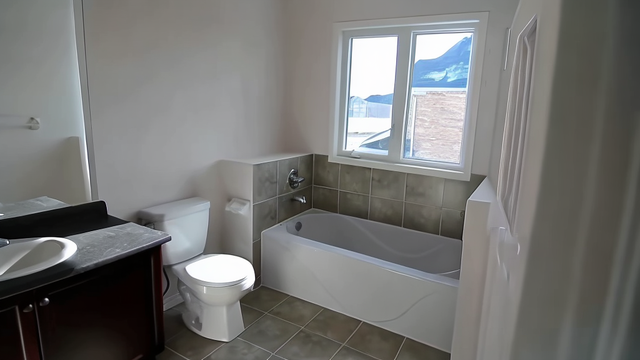} \\[1pt]
        \includegraphics[width=\vsw]{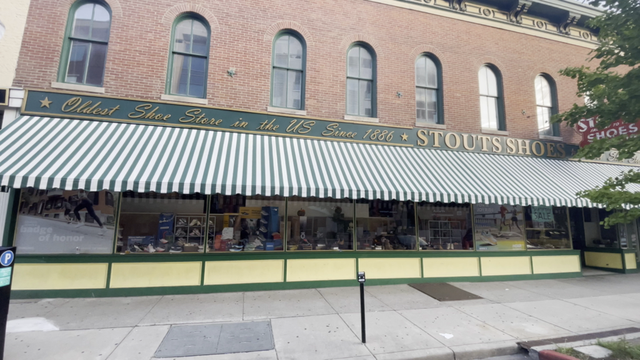} &
        \includegraphics[width=\vsw]{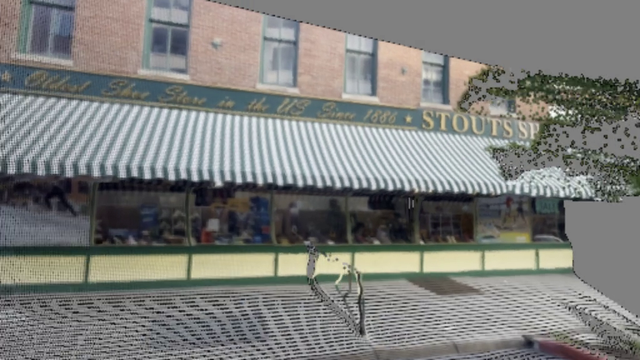} &
        \includegraphics[width=\vsw]{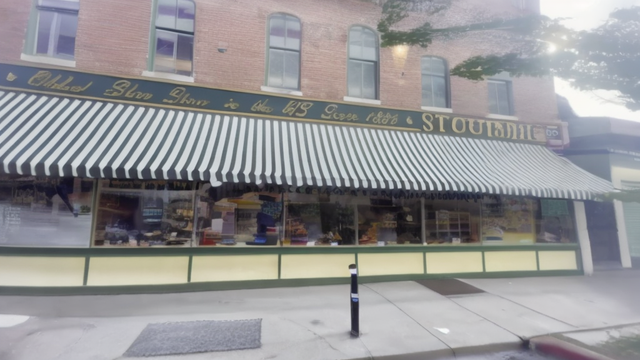} \\[1pt]
        \includegraphics[width=\vsw]{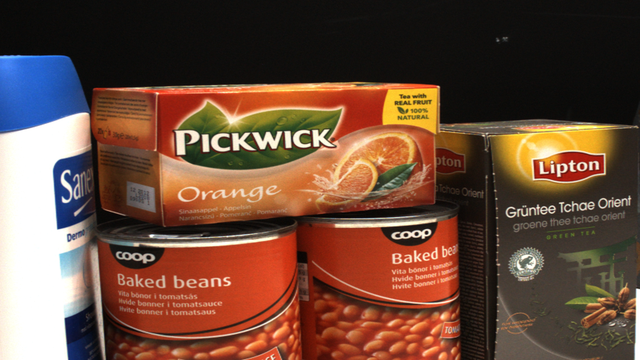} &
        \includegraphics[width=\vsw]{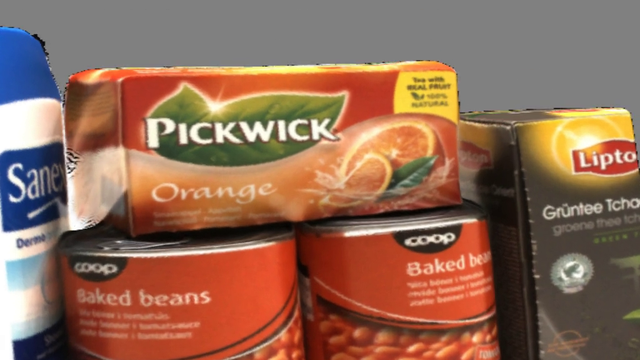} &
        \includegraphics[width=\vsw]{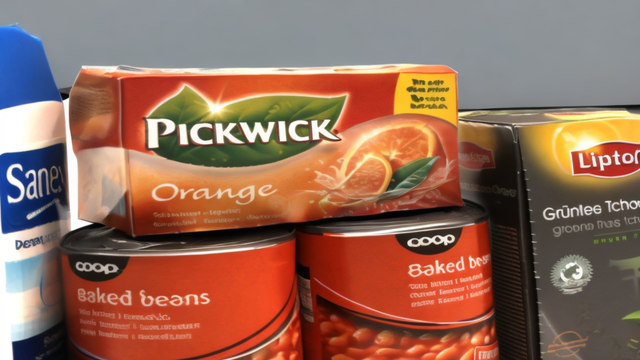} \\[1pt]
        \includegraphics[width=\vsw]{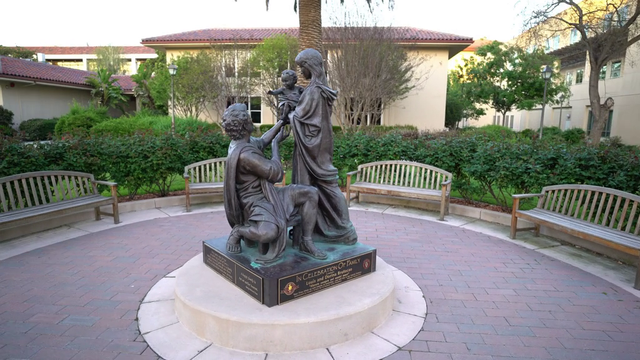} &
        \includegraphics[width=\vsw]{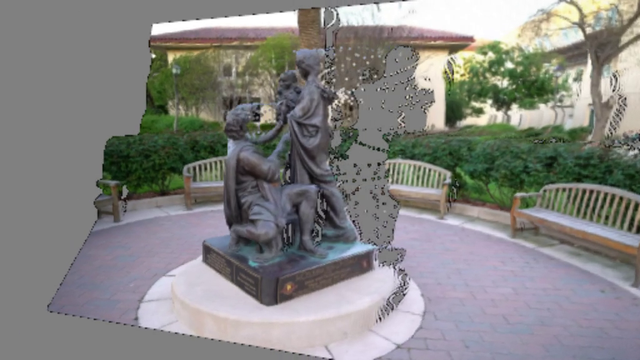} &
        \includegraphics[width=\vsw]{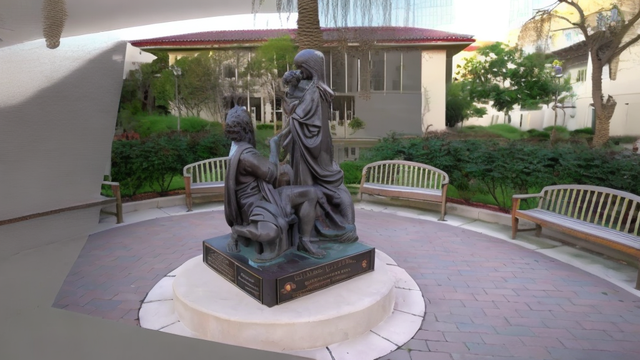} \\[2pt]
        {\scriptsize GT} & {\scriptsize \makecell{Point-cloud raster \\ (conditioning)}} & {\scriptsize \makecell{\viewcrafter \\ (diffusion)}} \\
    \end{tabular}
    \caption{
        {\bf \viewcrafter\ point-cloud conditioning vs.\ diffusion output. }
        \viewcrafter\ rasterizes its single-image point cloud from the (canonical) target pose to form the conditioning frame (middle), then a video-diffusion model synthesizes the final view (right).
        The raster is geometrically and photometrically faithful wherever the single view provides coverage (the speckled regions are sparse splatting where coverage thins); the diffusion model fills the remainder but can hallucinate geometry and drift in color as the viewpoint rotates away from the input.
        Rows: RealEstate10K (SV), RealEstate10K (VC), DL3DV-10K, DTU, Tanks \& Temples (VC), on the same scenes as \cref{fig:qualitative}.
    }
    \label{fig:viewcrafter_stages}
\end{figure}